\documentclass[10pt,journal,compsoc]{IEEEtran}
\ifCLASSOPTIONcompsoc
  \usepackage[nocompress]{cite}
\else
  \usepackage{cite}
\fi
\ifCLASSINFOpdf
\else
\fi
\usepackage{amsmath}
\usepackage{multirow}
\usepackage{graphicx}
\usepackage{amsfonts}
\usepackage{amsthm}
\usepackage[utf8]{inputenc}

\newtheorem{theorem}{Theorem}
\theoremstyle{definition}
\newtheorem{definition}{Definition}[section]
\newtheorem{proposition}[theorem]{Proposition}

\usepackage{booktabs}
\usepackage{algorithm}
\usepackage{algorithmicx}
\usepackage{algpseudocode}

\usepackage{makecell}

\usepackage{url}

\begin{document}
%
\title{Recognizing Predictive Substructures with Subgraph Information Bottleneck}
%
%
%
%

\author{Junchi~Yu,
        Tingyang~Xu,
        Yu~Rong,
        Yatao~Bian,
        Junzhou~Huang,
        and~Ran~He,~\IEEEmembership{Senior Member,~IEEE}
\IEEEcompsocitemizethanks{\IEEEcompsocthanksitem Junchi Yu and Ran He are with the Center for Research on Intelligent Perception and
Computing (CRIPAC), National Laboratory of Pattern Recognition, Institute of Automation, Chinese Academy of Sciences, Beijing, China 
. Ran He is also with the Center for Excellence in Brain Science and Intelligence Technology, Chinese Academy of Sciences, Beijing, China.\protect\\
E-mail: yujunchi2019@ia.ac.cn, rhe@nlpr.ia.ac.cn.
\IEEEcompsocthanksitem Tingyang Xu, Yu Rong, Yatao Bian and Junzhou Huang are with the Tencent AI LAB, Shenzhen, China.\protect\\
E-mail: tingyangxu, yu.rong, yataobian, joehuang@tencent.com.
\IEEEcompsocthanksitem Ran He is the corresponding author.}
\thanks{Manuscript received April 19, 2005; revised August 26, 2015.}}

%
%

\markboth{Journal of \LaTeX\ Class Files,~Vol.~14, No.~8, August~2015}%
{Shell \MakeLowercase{\textit{et al.}}: Bare Demo of IEEEtran.cls for Computer Society Journals}
%



\IEEEtitleabstractindextext{%
\begin{abstract}

The emergence of Graph Convolutional Network (GCN) has greatly boosted the progress of graph learning. However, two disturbing factors, noise and redundancy in graph data, and lack of interpretation for prediction results, impede further development of GCN. One solution is to recognize a predictive yet compressed subgraph to get rid of the noise and redundancy and obtain the interpretable part of the graph. This setting of subgraph is similar to the information bottleneck (IB) principle, which is less studied on graph-structured data and GCN. Inspired by the IB principle, we propose a novel subgraph information bottleneck (SIB) framework to recognize such subgraphs, named IB-subgraph. However, the intractability of mutual information and the discrete nature of graph data makes the objective of SIB notoriously hard to optimize. To this end, we introduce a bilevel optimization scheme coupled with a mutual information estimator for irregular graphs. Moreover, we propose a continuous relaxation for subgraph selection with a connectivity loss for stabilization. We further theoretically prove the error bound of our estimation scheme for mutual information and the noise-invariant nature of IB-subgraph. Extensive experiments on graph learning and large-scale point cloud tasks demonstrate the superior property of IB-subgraph.

\end{abstract}

\begin{IEEEkeywords}
Graph Convolutional Network, Subgraph Information Bottleneck, Graph Classification.
\end{IEEEkeywords}}

\maketitle

\IEEEdisplaynontitleabstractindextext

%
\IEEEpeerreviewmaketitle

\IEEEraisesectionheading{\section{Introduction}\label{sec:introduction}}

%
%
%
%

\IEEEPARstart{M}{any} real-world data such as social networks, drug molecules, and point clouds can be viewed as graphs. Classifying the labels or analyzing the underlying properties of graph-structured data is the fundamental problem in the deep graph learning area. Recently, there is a surge of interest to employ Graph Convolutional Network (GCN) \cite{gcn} to cope with such irregular graphs. Many GCN variants \cite{diffpool,asapool,specpool} have achieved new state-of-the-art performance in various graph-level tasks. Apart from the tremendous progress in deep graph learning, the main concern is that real-world graphs are likely to contain redundant even noisy structure information \cite{conf/icml/FranceschiNPH19, journals/corr/abs-1911-07123}. This is detrimental to various GCN variants as they are incapable of recognizing the poisoned structure \cite{conf/iclr/ZugnerG19, dai2018adversarial,chang2020restricted}. Moreover, since GCN variants generally leverage all structure information for prediction \cite{diffpool, velickovic2017graph}, the interpretation for their prediction results has lagged compared to the significant gain in performance. These two challenges for the literature of graph learning trigger an interesting idea to recognize a subgraph which is predictive in terms of the graph label and discards unnecessary information.




Recognizing the predictive yet compressed subgraph sheds lights on various tasks as it discovers the vital substructure and filters out irrelevant parts.
For example, in drug discovery, when viewing molecules as graphs with atoms as nodes and chemical bonds as edges, biochemists are interested in identifying the subgraphs that mostly represent certain properties of the molecules, namely the functional groups \cite{substructure,mmpn}. In graph representation learning, the predictive subgraph highlights the vital substructure for graph classification, and provides an alternative way for yielding graph representation besides mean/sum aggregation \cite{gcn,velickovic2017graph,Xu:2019ty} and pooling aggregation \cite{diffpool,sapool,specpool}. In graph attack and defense, it is vital to purify a perturbed graph and mine the robust structures for classification \cite{journals/corr/abs-2005-10203}. In 3D object detection and segmentation, viewing point clouds as K-NN graphs, researchers seek for substructures which share the same categories \cite{chen2020hierarchical,lei2020seggcn,lin2020convolution,shi2020point,xu2020grid}.

Along with wide applications, the major difficulty is that it is laborious and time-consuming to obtain explicitly subgraph-level annotations for supervised learning. For example, there are 250 thousand molecules in ZINC250K dataset \cite{irwin2005zinc}. It requires extra expertise and experience to label every functional group in each molecule. In the absence of explicit supervision, one is blind to recognize the predictive yet compressed subgraph. Recently, the mechanism of self-attentive aggregation \cite{hierG,conf/nips/KnyazevTA19} employs attention scores to validate the significance of nodes and somehow discovers a vital substructure at node level with a well-selected threshold. However, the discovered substructure is highly influenced by the threshold. Moreover, this method only identifies isolated important nodes but ignores the topological information at the subgraph level. Consequently, it leads to a novel challenge as subgraph recognition: 
\emph{how can we recognize a compressed subgraph with minimum information loss in terms of predicting the graph labels/properties?}

Recalling the above challenge, there is a similar problem setting in information theory called information bottleneck (IB) principle \cite{ib}, which aims to juice out a compressed code from the original data that keeps most predictive information of labels or properties. Recently, there is a growing tendency to incorporate IB principle with deep learning due to its capability of extracting informative representation from regular data in the fields of computer vision \cite{rl_ib1,vib,conf/iccv/LuoLGYY19}, reinforcement learning \cite{rl_ib2,rl_ib3} and natural language processing \cite{nlp_ib1}. However, current IB methods, like VIB \cite{vib}, are still incapable for irregular graph data. Another work, Graph Information Bottleneck \cite{wu2020graph}, attempts to bind IB with node representation learning. It follows the Gaussian prior assumption in \cite{vib}, iteratively sample neighbors and learning node representation at each layer of the network, which is still far from subgraph recognition. Hence, it is still challenging for IB to compress irregular graph data, like a subgraph from an original graph, with a minimum information loss. 


To this end, we advance the IB principle for irregular graph data to solve the proposed subgraph recognition problem, which leads to a novel principle, Subgraph Information Bottleneck (SIB). SIB directly recognizes a compressed subgraph which is most predictive of a certain graph label without any subgraph annotations. This idea significantly distinguishes from prior researches in IB in two aspects. First of all, SIB directly reveals the vital substructure at the subgraph level instead of learning an optimal representation of the input data in the hidden space. Secondly, SIB deals with discrete and irregular graph data rather than regular data. We first i) leverage the mutual information estimator from Deep Variational Information Bottleneck (VIB) \cite{vib} for irregular graph data as the SIB objective. 
However, VIB is intractable to compute the mutual information without knowing the distribution forms, especially on graph data. 
To tackle this issue, ii) we adopt a bi-level optimization scheme to maximize the SIB objective. 
Meanwhile, the continuous relaxation that we adopt to approach the discrete selection of subgraph will lead to unstable optimization process. 
To further stabilize the training process and encourage a compact subgraph, iii) we propose a novel connectivity loss to assist SIB to effectively discover the maximally informative but compressed subgraph, which is defined as IB-Subgraph. 
By optimizing the above SIB objective and connectivity loss, one can recognize the IB-Subgraph only with the input graph and its label/property.
On the other hand, iv) SIB is model-agnostic and can be easily plugged into various Graph Neural Networks (GNNs).

We evaluate SIB in four application scenarios: improvement of graph classification, graph interpretation, graph denoising, and 3D relevant structure extraction.  Extensive experiments on both synthetic and real-world datasets demonstrate that the information-theoretic IB-Subgraph, recognized by the proposed Subgraph Information Bottleneck, enjoys superior graph properties compared to the subgraphs found by SOTA baselines. 

This paper is an extension of the conference work \cite{yu2020graph}. There are three major improvements over the previous one:
\textbf{1). We theoretically analyze with provable guarantees that the IB-subgraph, found by SIB framework, is noise-invariant.} In the previous version, we empirically analyze that the significant performance gain in various tasks is because SIB is able to preserve the informative substructure and discard noise and redundancy. In this paper, we further indicate that optimizing SIB objective is equivalent to minimize the upper bound of the mutual information of the IB-subgraph and noise, and thus leads to the noise-invariance nature of IB-subgraph. \textbf{2). We provide more theoretical analysis on our optimization scheme.} We give the error bound when estimating the mutual information following the PAC-Bayes framework.  \textbf{3). We provide more experiment results on computer vision tasks.} Despite of graph interpretation, graph classification and graph denoising, we further evaluate the proposed method on 
S3DIS point cloud dataset \cite{armeni_cvpr16}. Experiment results show the proposed method is compatible to the large GNN models and efficiently extracts label-relevant structures on large-scale graph datasets.

\section{Related Work}

\textbf{Subgraph Discovery.} Traditional subgraph discovery mainly includes dense subgraph discovery and frequent subgraph mining. Dense subgraph discovery aims to find the subgraph with the highest density (e.g. the number of edges over the number of nodes \cite{densesub,conf/kdd/GionisT15}). Frequent subgraph mining is to look for the most common substructure among graphs \cite{gspan,subdue,SLEUTH}. Recently, it is popular to select a neighborhood subgraph of a central node to do message passing in node representation learning. DropEdge \cite{rong2020dropedge} relieves the over-smoothing phenomenon in deep GCNs by randomly dropping a portion of edges in graph data. Similar to DropEdge, DropNode \cite{chen2018fastgcn,conf/nips/HamiltonYL17,huang2018adaptive} principle is also widely adopted in node representation learning.  FastGCN \cite{chen2018fastgcn} and ASGCN \cite{huang2018adaptive} accelerate GCN training via node sampling. GraphSAGE \cite{conf/nips/HamiltonYL17} leverages neighborhood sampling for inductive node representation learning. NeuralSparse \cite{zheng2020robust} select Top-K (K is a hyper-parameter) task-relevant 1-hop neighbors of a central node for robust node classification.  Similarly, researchers discover the vital substructure at node level via the attention mechanism \cite{velickovic2017graph,sapool,conf/nips/KnyazevTA19}. \cite{gnnexplainer} further identifies the important computational graph for node classification. \cite{sgnn} discovers subgraph representations with specific topology given subgraph-level annotation. However, the above methods are far from subgraph recognition, as they are incapable of discovering a compressed yet predictive subgraph in graph data.

\textbf{Graph Classification.} In recent literature, there is a surge of interest in adopting graph neural networks (GNN) in graph classification. The core idea  is to aggregate all the  node information for graph representation. A typical implementation  is the mean/sum aggregation \cite{gcn,Xu:2019ty}, which is to average or sum up the node embeddings. An alternative way is to leverage the hierarchical structure of graphs, which leads to the pooling aggregation \cite{diffpool,sortpool,sapool,specpool}.  When tackling redundant and noisy graphs, these approaches will likely to result in the sub-optimal graph representation. Recently, InfoGraph \cite{sun2019infograph} maximize the mutual information between graph representations and multi-level local representations to obtain more informative global representations.


\textbf{Information Bottleneck.} Information bottleneck (IB), originally proposed for signal processing, attempts to find a short code of the input signal but preserves maximum information of the code \cite{ib}. \cite{vib} firstly bridges the gap between IB and deep learning, and proposed variational information bottleneck (VIB). Nowadays, IB and VIB have been wildly employed in computer vision \cite{rl_ib1,conf/iccv/LuoLGYY19}, reinforcement learning \cite{rl_ib2,rl_ib3}, natural language processing \cite{nlp_ib1}, and speech and acoustics \cite{speech_ib1} due to the capability of learning compact and meaningful representations. However, IB is less researched on irregular graphs due to the intractability of mutual information. A parallel work, named Graph Information Bottleneck, recently incorporates IB with node representation learning. However, it is still far from directly recognizing a compressed but informative subgraph in graph data.

\textbf{Point Cloud Segmentation.} Point cloud segmentation is to identify the category of each element in the set of points, a type of 3D geometric data. Traditional convolutional network can not consume such irregular data. PointNet \cite{qi2017pointnet} first proposes a unified framework to process the point cloud. Recent breakthroughs in graph learning innovate researchers to view the point cloud as a K-NN graph, thus naturally leading to GCN-based solutions \cite{chen2020hierarchical,lei2020seggcn,lin2020convolution,shi2020point,xu2020grid, li2019deepgcns}. 
Beyond that, as it is expensive to label the point cloud, \cite{xu2020weakly, shi2021label} leverage a tiny portion of labeled point for weakly supervised training. Here, in the subgraph recognition phenomenon, one is required to infer the underlying substructure of a certain category with no point-level attribution, which is a huge challenge for existing methods.

\section{Notations and Preliminaries}

Let $\{(G_1, Y_1),\dots,(G_N, Y_N)\}$ be a set of $N$ graphs with their real value properties or categories, where $G_n$ refers to the $n$-th graph and $Y_n$ refers to the corresponding properties or labels. We denote by $G_n=(\mathbb{V},\mathbb{E}, A, X)$ the $n$-th graph of size $M_n$ with node set $\mathbb{V}=\{V_i|i=1,\dots,M_n\}$, edge set $\mathbb{E}=\{(V_i, V_j)|i>j; V_i,V_j \text{ is connected}\}$,
adjacent matrix $A\in \{0,1\}^{M_n\times M_n}$, and feature matrix $X\in R^{M_n\times d}$ of $V$ with $d$ dimensions, respectively. Denote the neighborhood of $V_i$ as $\mathcal{N}(V_i)=\{V_j|(V_i, V_j)\in \mathbb{E}\}$. 
We 
use 
$G_{sub}$ as a specific subgraph and $\overline{G}_{sub}$ as the complementary structure of $G_{sub}$ in $G$. Let $f:G \rightarrow R / [0,1,\cdots,n] $ be the mapping from graphs to the real value property or category, $Y$. $\mathbb{G}$ is the domain of graphs. $I(X,Y)$ refers to the Shannon mutual information of two random variables.

\subsection{Graph convolutional network}

Graph convolutional network (GCN) is widely adopted to graph classification.  Given a graph $G = (V,\mathbb{E})$ with node feature $X$ and adjacent matrix $A$, GCN outputs the node embeddings $X^{'}$ from the following process:
\begin{equation}
\begin{aligned}
X^{'} = \mathrm{GCN}(A,X;W) = \mathrm{ReLU}(D^{-\frac{1}{2}}\hat{A}D^{-\frac{1}{2}}XW), 
\end{aligned}
\end{equation}
where $D$ refers to the diagonal matrix with nodes' degrees and $\hat{A}=A+I$ is the adjacent matrix after adding the self-loop. $W$ refers to the model parameters.

One can simply sum up the node embeddings to get a fixed-length graph embeddings \cite{Xu:2019ty}. Recently, researchers attempt to exploit an hierarchical structure of graphs, which leads to various graph pooling methods \cite{hierG,conf/icml/GaoJ19,sapool,journals/corr/abs-1905-10990,sortpool,asapool,diffpool}. \cite{hierG} enhances the graph pooling with self-attention mechanism to leverage the importance of different nodes contributing to the results. Finally, the graph embedding is obtained by multiplying the node embeddings with the normalized attention scores:
\begin{equation}
\begin{aligned}
E = \mathrm{Att}(X^{'}) =\mathrm{ softmax}(\Phi_{2}\mathrm{tanh}(\Phi_{1}X^{'T}))X^{'},
\end{aligned}
\end{equation}
where $\Phi_1$ and $\Phi_2$ refer to the model parameters of self-attention.

\subsection{Information Bottleneck}

Given the input data $X$ and the label $Y$, the information bottleneck principle aims to discover the latent representation $Z$ which is maximally informative in terms of $Y$ (\textbf{Sufficient}) and contains as little information of the input data $X$ as possible (\textbf{Minimal}). Formally, the sufficient and minimal representation, denoted as $Z_{s}$ and $Z_{m}$, can be obtained by the following objectives respectively:

\begin{equation}
\begin{aligned}
Z_{s} &= \arg\max\limits_{Z}I(Z,Y)\\
Z_{m} &= \arg\min\limits_{Z}I(Z,X)
\end{aligned}
\end{equation}
where $I(A,B)$ is the mutual information between random variable $A$ and $B$:
\begin{equation}
\begin{aligned}
I(A,B) = \int_{a\in A}\int_{b\in B} p(a,b)\log\frac{p(a,b)}{p(a)p(b)} dadb
\end{aligned}
\end{equation}
Built upon the above intuition, one can learn the minimally sufficient $Z$ by maximizing the information bottleneck objective: 
\begin{equation}
\begin{aligned}
\mathcal{L}_{IB} = {I(Z,Y)-\beta I(X,Z)}
\end{aligned}
\end{equation}
where $\beta$ refers to a hyper-parameter trading off informativeness and compression. Optimizing this objective will lead to a minimally sufficient $Z$, which is less prone to over-fitting and less sensitive to noise. However, the IB objective is notoriously difficult to optimize as it is troublesome to compute the mutual information.  \cite{vib} optimize a tractable lower bound of the IB objective: 
\begin{equation}
\begin{aligned}
\mathcal{L}_{VIB} &= \frac{1}{N} \sum\nolimits_{i=1}^{N} \int\nolimits p(z|x_{i})\log{q_{\phi}(y_{i}|z)} dz\\
&- \beta \mathrm{KL}(p(z|x_{i})|r(z)),
\label{vib_obj}
\end{aligned}
\end{equation}
where $q_{\phi}(y_{i}|z)$ is the variational approximation to $p_{\phi}(y_{i}|z)$ and $r(z)$ is the prior distribution of $Z$. $\mathrm{KL}$ is the Kullback-Leibler divergence.
However, it is hard to estimate the mutual information in high dimensional space when the distribution forms are inaccessible, especially for irregular graph data.

\begin{figure*}[t]
\vspace{+0.2cm}
\begin{center}
\centerline{\includegraphics[width=1.8\columnwidth]{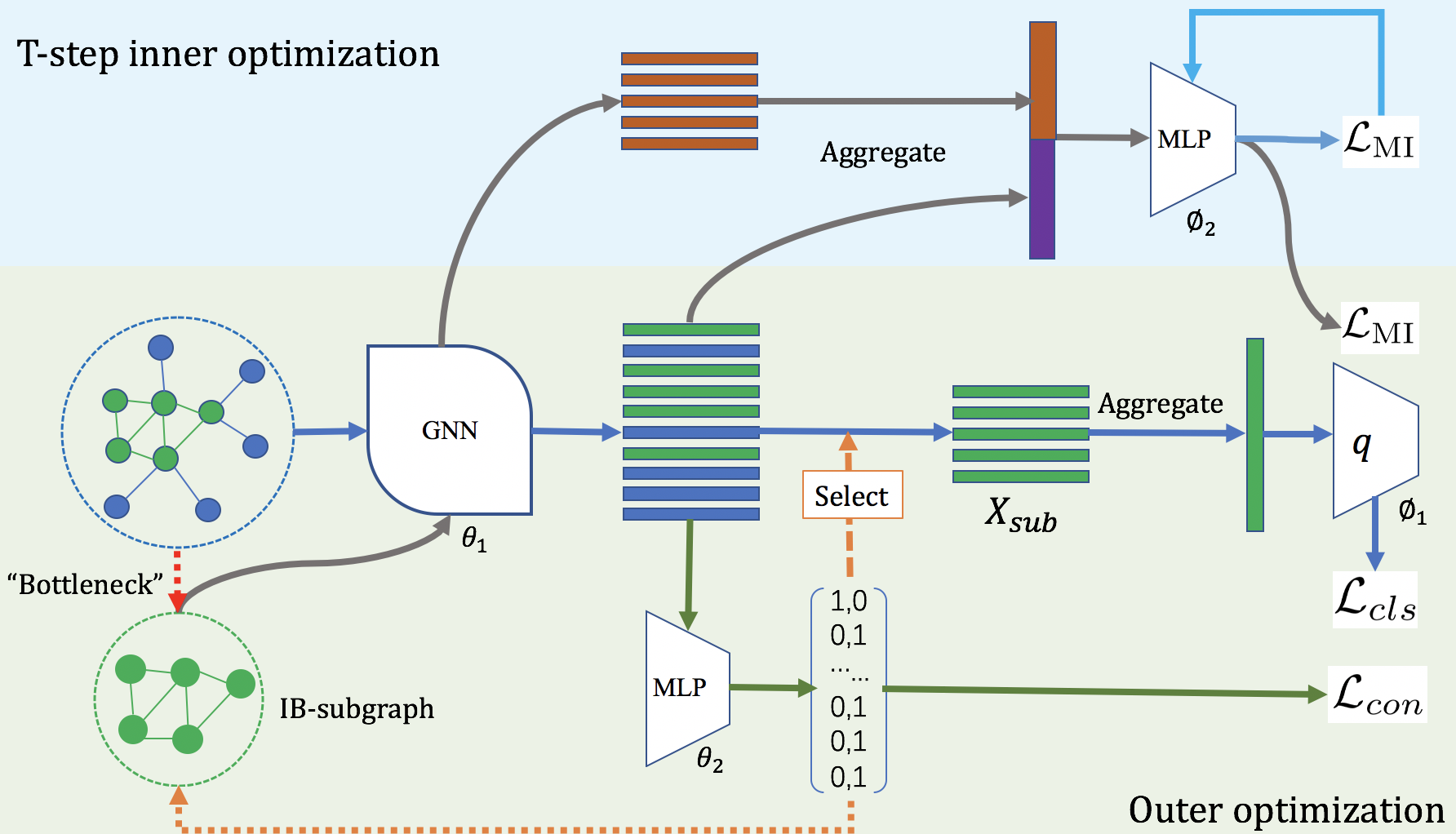}}
\end{center}
\vspace{+0.2cm}
\caption{Illustration of the  proposed subgraph information  bottleneck (SIB) framework. We employ a bi-level optimization scheme to optimize the SIB objective and thus yielding the IB-Subgraph. In the  inner optimization phase, we estimate $I(G,G_{sub})$ by optimizing the statistics network of the DONSKER-VARADHAN representation \cite{dv-representation}.  Given a good estimation of $I(G,G_{sub})$, in the outer optimization phase, we maximize the SIB objective by optimizing the mutual information,  the classification loss $\mathcal{L}_{cls}$ and connectivity loss $\mathcal{L}_{con}$. } 
\label{flowchart}
\end{figure*}

\section{Optimizing the Subgraph Information Bottleneck Objective for Subgraph Recognition}

In this section, we will elaborate on the proposed method in detail. We first formally define the subgraph information bottleneck and IB-Subgraph. Then, we introduce a novel framework for SIB  to effectively find the IB-Subgraph. Moreover, we propose a bi-level optimization scheme and a graph mutual information estimator for SIB optimization. We further elaborate a continuous relaxation strategy with a connectivity loss to stabilize the training.


\subsection{Subgraph Information Bottleneck}

To begin with, we formally define the \textbf{Minimal Subgraph} and the \textbf{Sufficient Subgraph}:

\begin{definition}
$G_{sub}$ is the \textbf{Minimal Subgraph} if it contains minimally usable information of $G$. We denote it as $G_{sub}^{m}=\arg\min_{G_{sub}}I(G, G_{sub})$.
\label{minimal-subgraph}
\end{definition}

Intuitively, the minimal subgraph compresses the information via dropping a portion of the topological structures of the corresponding graph. By the definition of mutual information, we have:

\begin{equation}
\begin{aligned}
I(G_{sub},G)&=H(G_{sub})-H(G_{sub}|G)\\
&\leq H(G_{sub})\leq\log|\mathbb{G}_{sub}|,\\
\end{aligned}
\label{natural}
\end{equation}
where $|\cdot|$ denotes the cardinally of a set. Eq.~\ref{natural} provides a natural bound of $I(G_{sub},G)$. That is, given a finite set $\mathbb{G}$, one can minimize the number of the elements in the induced subgraph set $\mathbb{G}_{sub}$. One possible way is to discover the most common and frequent substructures among graphs \cite{gspan,subdue,SLEUTH}. However, it is troublesome to directly optimize the above bound since it is quite loose and hard to optimize with gradient methods. 
 
\begin{definition}
$G_{sub}$ is the \textbf{Sufficient Subgraph} if it maximizes the mutual information $I(G_{sub},Y)$, where $Y$ is the label of the original graph $G$. We denote it as $G_{sub}^{s}=\arg\max_{G_{sub}}I(Y, G_{sub})$.
\label{suff-subgraph}
\end{definition}

A sufficient subgraph ensures that almost all substructures which are essential to the graph property are preserved. Moreover, when defining a Markov Chain $Y\rightarrow G\rightarrow G_{sub}$, we have $I(Y, G_{sub})\leq I(G,Y)$ by the data processing inequality. This leads to $I(G_{sub}^{s},Y)=I(G,Y)$.  As a consequence, we can achieve comparable performance in the downstream tasks when replacing $G$ with $G_{sub}$. However, there is a trivial solution where $G_{sub}^{s}=G$. In this case, we are unable to distinguish the essential subgraphs and the noise and redundancy.


To this end, we generalize the information bottleneck principle to recognize minimal sufficient subgraphs in irregular graphs, which leads to the subgraph information bottleneck (SIB) principle. 

\begin{definition}[Subgraph Information Bottleneck]
Given a graph $G$ and its label $Y$, the Subgraph Information Bottleneck seeks for the most informative yet compressed subgraph $G_{sub}$ by optimizing the following objective:

\begin{equation}
\begin{aligned}
&\max\limits_{G_{sub}} I(Y,G_{sub})-\beta I(G,G_{sub}).
\end{aligned}
\label{gib-sub}
\end{equation}

\end{definition}

We denote the subgraph $G_{sub}$ induced by Eq.~\ref{gib-sub} as the \textbf{IB-Subgraph}. As is shown in Eq.~\ref{gib-sub}, learning the IB-subgraph needs no subgraph-level annotations. One is supposed to obtain the IB-subgraph only with the input graphs and their labels, thanks to the information-theoretic SIB objective. Statistically, IB-Subgraph is minimal sufficient concerning the original graph. Intuitively, the IB-Subgraphs only preserve label-relevant substructure in original graphs, and thus reduce the effect of noise and redundancy to the downstream tasks. Specifically, let $G_{n}$ be the noisy substructure which is irrelevant to the graph property $Y$, we show the IB-Subgraph $G_{sub}$ is invariant to the noise $G_{n}$.  


\begin{proposition}[Noise-invariance]
\label{noise-invariance}
Suppose the noisy structure $G_{n}$ is independent of $Y$, the mutual information $I(G_{n}, G_{sub})$ is upper bounded by $I(G_{sub}, G)-I(G_{sub},Y)$:

\begin{equation}
\begin{aligned}
I(G_{n}, G_{sub})\leq I(G,G_{sub})-I(Y,G_{sub}).
\end{aligned}
\end{equation}

\end{proposition}
\begin{proof}
We prove the above proposition following the Markov chain assumption in \cite{achille2018emergence}. Suppose $G_{n}$ interacts with $G_{sub}$ only through $G$ and $G$ is defined by $Y$ and $G_{n}$. We can define the following Markov chain $(Y, G_{n})\rightarrow G\rightarrow G_{sub}$. By the Data Processing Inequality, we obtain:
\begin{equation}
\begin{aligned}
I(G;G_{sub})&\geq I(Y,G_{n}; G_{sub}) \\
& = I(G_{n};G_{sub}) + I(Y;G_{sub}|G_{n}) \\
& = I(G_{n};G_{sub}) + H(Y|G_{n}) - H(Y|G_{n};G_{sub}), \\
\end{aligned}
\label{invariant}
\end{equation}

since $G_{n}$ and $Y$ are independent, we obtain $H(Y|G_{n})=H(Y)$. Moreover, it is straightforward that $H(Y|G_{n};G_{sub})\leq H(Y|G_{sub})$. Plug the above equality and inequality into Eq.~\ref{invariant}, and we derive:
\begin{equation}
\begin{aligned}
I(G;G_{sub})&\geq I(G_{n};G_{sub}) + H(Y) - H(Y|G_{sub})\\
& = I(G_{n};G_{sub}) + I(Y;G_{sub}),
\end{aligned}
\end{equation}
thus we obtain $I(G_{n}, G_{sub})\leq I(G,G_{sub})-I(Y,G_{sub})$ and conclude the proof.
\end{proof}

Proposition~\ref{noise-invariance} indicates that optimizing the SIB objective in Eq.~\ref{gib-sub} is equivalent to minimize the mutual information between the IB-Subgraph and the noisy substructure, leading to the noise-invariance property of the IB-Subgraph. This property is appealing as it provides theoretical guarantees that one can effectively discover the vital substructure which mostly influences the property of the original graph by compressing the information in the IB-Subgraph, even in the absence of explicit subgraph-level annotations. Meanwhile, as the SIB objective also restrict the IB-Subgraph to be predictive, it is sufficient to plug SIB objective into various baseline models to enhance their performances, which shows SIB is model-agnostic.



However, the SIB objective in Eq.~\ref{gib-sub} is notoriously hard to optimize due to the intractability of mutual information and the discrete nature of irregular graph data. We then introduce approaches on  how to optimize such objective.

\subsection{Bi-level optimization for the SIB objective}

The SIB objective in Eq.~\ref{gib-sub} consists of two parts. We examine the first term $I(Y,G_{sub})$ in Eq.~\ref{gib-sub}. This term measures the relevance between $G_{sub}$ and $Y$: 

\begin{equation}
\begin{aligned}
I(Y,G_{sub}) = \int p(Y,G_{sub}) \log{{p(Y|G_{sub})}} dyY\  dG_{sub} + \mathrm{H}(Y). \label{sub-obj}\\
\end{aligned}
\end{equation}
$H(Y)$ is the entropy of $Y$ and thus can be ignored. In practice, we approximate $p(Y,G_{sub})$ with an empirical distribution $p(Y,G_{sub}) \approx\frac{1}{N} \sum_{i=1}^{N}\delta_{Y}(y_{i})\delta_{G_{sub}}(G_{sub_{i}})$, where $\delta()$ is the Dirac function to sample training data. $G_{sub_{i}}$ and $Y_{i}$ are the output subgraph and graph label corresponding to i-th training data. By substituting the true posterior $p(Y|G_{sub})$ with a variational approximation $q_{\phi_{1}}(Y|G_{sub})$, we obtain a tractable lower bound of the first term in Eq.~\ref{gib-sub}:
\begin{equation}
\begin{aligned}
I(Y,G_{sub}) &= H(Y|G_{sub}) + H(Y)  \\
&= \int p(Y,G_{sub}) \log{{q_{\phi_{1}}(Y|G_{sub})}} dY\\
&+ \mathrm{KL}[p(Y|G_{sub})|q_{\phi_{1}}(Y|G_{sub}))]\\
&\geq \int p(Y,G_{sub}) \log{{q_{\phi_{1}}(Y|G_{sub})}} dY \ dG_{sub} \\
&\approx \frac{1}{N} \sum_{i=1}^{N} \log{q_{\phi_{1}}(Y_{i}|G_{sub_{i}}}) \\
&=: -\mathcal{L}_{cls}(q_{\phi_{1}}(Y|G_{sub}),Y_{gt}), \\
\end{aligned}
\label{mi_1}
\end{equation}

where $Y_{gt}$ is the ground truth label of the graph. Eq.~\ref{mi_1} indicates that maximizing $I(Y,G_{sub})$ is achieved by the minimization of the classification loss between $Y$ and $G_{sub}$ as  $\mathcal{L}_{cls}$. Intuitively, minimizing $\mathcal{L}_{cls}$ encourages the subgraph to be predictive of the graph label, and thus leads to the relevance between $G_{sub}$ and $Y$. In practice, we choose the cross entropy loss for categorical $Y$ and the mean squared loss for continuous $Y$, respectively. Then we introduce the PAC-Bayes bound:

\begin{theorem}
\label{proof-theorem}

Suppose $G_{sub}$ and $Y$ take value in $\mathcal{G}$ and $\mathcal{Y}$ respectively. Let $f(Y,G_{sub}) = \log{q_{\phi_{1}}(Y|G_{sub})}\in [-B,B]$, and $\mathcal{F}$ is the family of $f(Y,G_{sub})$. let $(\mathcal{F},d)$ be a metric space, $d=||\cdot||_{2}$. Define $\mathcal{F}_{t}=\{(f_{1},f_{2},\cdots,f_{n})|\forall{f_{k}}\in \mathcal{F},\exists{i}\in 1\cdots,n,d(f_{k},f_{i})\leq t\}$ is a t-cover of $\mathcal{F}$, and $\mathcal{N}(t,\mathcal{F},d)$ is the covering number of $\mathcal{F}$ . Then, for any $\delta \in (0,0.5)$, with probability $p\geq 1-\delta$ we have the following inequality
,  up to a constant:
\begin{equation}
\begin{aligned}
&|I(Y,G_{sub})-\frac{1}{N} \sum_{i=1}^{N} \log{q_{\phi_{1}}(Y_{i}|G_{sub_{i}}})| \\
&\leq B\sqrt{\frac{2\log{\frac{1}{\delta}}}{N}} + \inf_{t}2(t+\sqrt{\frac{\log{\mathcal{N}(t,\mathcal{F},d)}}{N}}) \\
&+ \mathrm{KL}[p(Y|G_{sub})|q_{\phi_{1}}(Y|G_{sub}))]. \\
\end{aligned}
\end{equation}

\end{theorem}


Theorem~\ref{proof-theorem} indicates that we can reduce the estimating error with large training sets and simple prediction models. Moreover, the $\mathrm{KL}$ term decreases when the variational approximation approaches $p(Y|G_{sub})$. This also reduces the estimating error. The proof is in supplementary materials.  Then, we consider the minimization of $I(G,G_{sub})$ which is the second term of Eq.~\ref{gib-sub}. Remind that \cite{vib} introduces a tractable prior distribution $r(Z)$ in Eq.~\ref{vib_obj}, and thus results in a variational upper bound. However, this setting is troublesome as it is hard to find a reasonable prior distribution for $p(G_{sub})$, which is the distribution of graph substructures instead of latent representation. Thus we go for another route.  Directly applying the DONSKER-VARADHAN representation \cite{dv-representation} of the KL-divergence, we have:
\begin{equation}
\begin{aligned}
I(G,G_{sub}) =& \sup \limits_{f_{\phi_2}:\mathbb{G}\times \mathbb{G}\rightarrow \mathbb{R}} \mathbb{E}_{G,G_{sub}\in p(G,G_{sub})}f_{\phi_{2}}(G,G_{sub})\\
&-\log{\mathbb{E}_{G \in p(G),G_{sub}\in p(G_{sub})}e^{f_{\phi_{2}}(G,G_{sub})}},
\end{aligned}
\label{mi_est}
\end{equation}
where $f_{\phi_2}$ is the statistics network that maps from the graph set to the set of real numbers. 
In order to approximate $I(G,G_{sub})$ using Eq.~\ref{mi_est}, we design a statistics network  based on modern GNN architectures as shown by Figure \ref{flowchart}: first, we use a GNN to extract  embeddings from both $G$ and $G_{sub}$ (parameter shared with the subgraph generator, which will be elaborated in Section \ref{sec_subgraph_generator}), then concatenate $G$ and $G_{sub}$ embeddings and feed them into a Multi-Layer Perceptron (MLP), which finally produces the real number.    
In conjunction with the sampling method to approximate $p(G,G_{sub})$, $p(G)$ and $p(G_{sub})$, we reach the following optimization problem to approximate\footnote{Notice that the MINE estimator \cite{mine} straightforwardly uses the DONSKER-VARADHAN representation to derive an MI estimator between the  regular input data and its vectorized representation/encoding. It cannot be applied to estimate the mutual information between $G$ and $G_{sub}$ since both $G$ and $G_{sub}$ are irregular graph data.} $I(G,G_{sub})$: 
\begin{equation}
\begin{aligned}
\max_{\phi_{2}} \quad \mathcal{L}_{\mathrm{MI}}(\phi_{2},G_{sub})& =  \frac{1}{N}\sum_{i=1}^{N}f_{\phi_{2}}(G_{i},G_{sub_{i}}) \\
&-\log{\frac{1}{N}\sum_{i=1,j\neq i}^{N}e^{f_{\phi_{2}}(G_{i},G_{sub_{j}})}}.
\end{aligned}
\label{mi_loss}
\end{equation}

    \begin{algorithm}[t]
        \caption{Optimizing the subgraph information bottleneck.}
        \label{pseudo-code}
        \begin{algorithmic}[1] 
            \Require Graph $G=\{A,X\}$, graph label $Y$, inner-step $T$, outer-step $N$. 
            \Ensure Subgraph $G_{sub}$
            \Function {SIB}{$G=\{A,X\}, Y, T, N$}
                \State $\theta \gets \theta^{0},\quad \phi_{1} \gets \phi_{1}^{0}$
                \For{$i=0\to N$}
                    \State $\phi_{2} \gets \phi_{2}^{0}$
                    \For{$t=0 \to T$}
                    
                        \State $\phi_{2}^{t+1} \gets \phi_{2}^{t} + \eta_{1} \nabla_{\phi_{2}^{t}}\mathcal{L}_{\rm MI} $
                    \EndFor
                        
                    \State $\theta^{i+1} \gets \theta^{i} -\eta_{2} \nabla_{\theta^{i}} \mathcal{L}(\theta^{i},\phi_{1}^{i},\phi_{2}^{T})  $
                    \State $\phi_{1}^{i+1} \gets \phi_{1}^{i} -\eta_{2} \nabla_{\phi_{1}^{i}} \mathcal{L}_(\theta^{i},\phi_{1}^{i},\phi_{2}^{T})  $
                \EndFor
                
                \State $G_{sub} \gets g(G;\theta^{N})$
                \State \Return{$G_{sub}$}
            \EndFunction
        \end{algorithmic}
    \end{algorithm}


With the approximation to the MI in graph data, we combine Eq.~\ref{gib-sub} , Eq.~\ref{mi_1} and Eq.~\ref{mi_loss}, and formulate the optimization process of SIB as a tractable bi-level optimization problem:
\begin{align}
&\min \limits_{G_{sub},\phi_{1}} \quad \mathcal{L}(G_{sub},\phi_{1},\phi_{2}^{*}) = \mathcal{L}_{cls} + \beta \mathcal{L}_{\rm MI} \label{bilevel}\\
&\text{ s.t. } \quad \phi_{2}^{*} = \mathop{\arg\max} \limits_{\phi_{2}}\mathcal{L}_{\mathrm{MI}} \label{t-steps}.
\end{align}
We first derive a sub-optimal $\phi_2$ notated as $\phi_2^*$ by optimizing Eq.~\ref{t-steps} for T steps as inner loops. After the T-step optimization of the inner-loop ends, Eq.~\ref{mi_loss} is a proxy for MI minimization for the SIB objective as an outer loop. Then, the parameter $\phi_{1}$ and the subgraph $G_{sub}$ are optimized to yield IB-Subgraph. 
However, in the outer loop, the discrete nature of $G$ and $G_{sub}$ hinders applying the gradient-based method to optimize the bi-level objective and find the IB-Subgraph.   

\begin{table*}[t]
\vspace{0.2cm}
  \centering
  \caption{Classification accuracy. The pooling methods yield pooling aggregation while the backbones yield mean aggregation. The proposed SIB method with backbones yields subgraph embedding by aggregating the nodes in subgraphs.}
    \begin{tabular}{ccccc}
    \toprule
    \textbf{Method} & \textbf{MUTAG} & \textbf{PROTEINS} & \textbf{IMDB-BINARY} & \textbf{DD} \\
    \midrule
     SortPool & \textbf{0.844 $\pm$ 0.141} & 0.747 $\pm$ 0.044 & 0.712 $\pm$ 0.047 & 0.732 $\pm$ 0.087 \\
           ASAPool  & 0.743 $\pm$ 0.077 & 0.721 $\pm$ 0.043 & 0.715 $\pm$ 0.044 & 0.717 $\pm$ 0.037 \\
           DiffPool & 0.839 $\pm$ 0.097 & 0.727 $\pm$ 0.046 & 0.709 $\pm$ 0.053 &  0.778 $\pm$ 0.030 \\
           EdgePool & 0.759 $\pm$ 0.077 & 0.723 $\pm$ 0.044 & 0.728 $\pm$ 0.044 & 0.736 $\pm$ 0.040 \\
           AttPool & 0.721 $\pm$ 0.086 & 0.728 $\pm$ 0.041 & 0.722 $\pm$ 0.047 & 0.711 $\pm$ 0.055 \\
    \midrule
     GCN   & 0.743$\pm$0.110 & 0.719$\pm$0.041 & 0.707 $\pm$ 0.037 & 0.725 $\pm$ 0.046 \\
           GraphSAGE & 0.743$\pm$0.077 & 0.721 $\pm$ 0.042 & 0.709 $\pm$ 0.041 & 0.729 $\pm$ 0.041 \\
           GIN   & 0.825$\pm$0.068 & 0.707 $\pm$ 0.056 & 0.732 $\pm$ 0.048 & 0.730 $\pm$ 0.033 \\
           GAT   & 0.738 $\pm$ 0.074 & 0.714 $\pm$ 0.040 & 0.713 $\pm$ 0.042 & 0.695 $\pm$ 0.045 \\
           GAT + DropEdge   & 0.743$\pm$0.081 &	0.711$\pm$0.043 &	0.710$\pm$0.041  &	0.717$\pm$0.035 \\           
    \midrule
     \textbf{GCN+SIB} & 0.776 $\pm$ 0.075 & 0.748 $\pm$ 0.046 & 0.722 $\pm$ 0.039 & 0.765 $\pm$  0.050 \\
     \textbf{GraphSAGE+SIB} & 0.760  $\pm$ 0.074 & 0.734 $\pm$ 0.043 & 0.719 $\pm$  0.052 & \textbf{0.781 $\pm$ 0.042} \\
      \textbf{GIN+SIB} & 0.839 $\pm$  0.064 & \textbf{0.749 $\pm$  0.051} & \textbf{0.737 $\pm$ 0.070} & 0.747 $\pm$  0.039 \\
      \textbf{GAT+SIB} & 0.749 $\pm$ 0.097 & 0.737 $\pm$ 0.044 & 0.729 $\pm$  0.046 & 0.769 $\pm$  0.040 \\
      \textbf{GAT+SIB+DropEdge} & 0.754$\pm$0.085&	0.737$\pm$0.037&	0.731$\pm$0.003&	0.776$\pm$0.034 \\      
    \bottomrule
    \end{tabular}%
  \label{tab:2}%
  \vspace{-0.0cm}
\end{table*}%

\begin{table}[t]
\vspace{0.2cm}
  \centering
  \caption{The mean and standard deviation of absolute property bias between the graphs and the corresponding subgraphs.}
  \setlength{\tabcolsep}{1mm}{
    \begin{tabular}{ccccc}
    \toprule
    \textbf{Method}    & \textbf{QED} & \textbf{DRD2}  & \textbf{HLM-CLint}   & \textbf{MLM-CLint} \\
    \midrule
    GCN+Att05 & 0.48$\pm$ 0.07 & 0.20$\pm$ 0.13 & 0.90$\pm$ 0.89 & 0.92$\pm$ 0.61 \\
    GCN+Att07 & 0.41$\pm$ 0.07 & 0.16$\pm$ 0.11 & 1.18$\pm$ 0.60 & 1.69$\pm$ 0.88 \\
    \textbf{GCN+SIB}  & \textbf{0.38$\pm$ 0.12} & \textbf{0.06$\pm$ 0.09} & \textbf{0.37$\pm$ 0.30} & \textbf{0.72$\pm$ 0.55} \\
    \bottomrule
    \end{tabular}}%
  \label{tab:1}%
  \vspace{0.0cm}
\end{table}%

\subsection{The Subgraph Generator}
\label{sec_subgraph_generator}

To alleviate the discreteness in Eq.~\ref{bilevel}, we propose the continuous relaxation to the subgraph recognition and propose a loss to stabilize the training process.

\textbf{Subgraph generator:} For the input graph $G$, we generate its IB-Subgraph with the node assignment $S$ which indicates the node is in $G_{sub}$ or $\overline{G_{sub}}$. Then, we introduce a continuous relaxation to the node assignment with the probability of nodes belonging to the $G_{sub}$ or $\overline{G_{sub}}$. For example, the $i$-th row of $S$ is a 2-dimensional vector $\textbf{[}p(V_{i}\in G_{sub}|V_{i}), p(V_{i} \in \overline{G_{sub}}|V_{i})\textbf{]}$. We first use an $l$-layer GNN to obtain the node embedding and employ a multi-layer perceptron (MLP) to output $S$ :

\begin{equation}
\begin{aligned}
X^{l}& = \mathrm{GNN}(A,X^{l-1};\theta_{1}), \\
S &= Softmax(\mathrm{MLP}(X^{l};\theta_{2})).
\end{aligned}
\end{equation}

$S$ is an $n\times 2$ matrix, where $n$ is the number of nodes. We add row-wise Softmax to the output of $\mathrm{MLP}$ to ensure the nodes are either in or out of the subgraph.  For simplicity, we compile the above modules as the subgraph generator, denoted as $g(;\theta)$ with $\theta:=(\theta_1, \theta_2)$. When $S$ is well-learned, the assignment of nodes is supposed to saturate to 0/1. The representation of $G_{sub}$, which is employed for predicting the graph label, can be obtained by taking the first row of $S^{T}X^{l}$.

\textbf{Connectivity loss:} When directly generating the IB-subgraph with the above subgraph generator, poor initialization will lead to the assignment $p(V_{i}\in G_{sub}|V_{i})$ and $p(V_{i}\in \overline{G_{sub}}|V_{i})$ of node to be close. Therefore, aggregation of the subgraph embedding by taking the first row of $S^{T}X^{l}$ is likely to contain excessive information of $\overline{G_{sub}}$. Moreover, as $S$ outputs the subgraph with a node selection manner, we hope our model to have an inductive bias to better leverage the topological information. That is, the found subgraph is supposed to be compact. Therefore, we propose the following connectivity loss:  
%
\begin{equation}
\begin{aligned}
\mathcal{L}_{con} = || \mathrm{Norm}(S^{T}AS)- I_2||_F,
\end{aligned}
\label{L_con}
\end{equation}
where $\mathrm{Norm(\cdot)}$ is the row-wise normalization, $||\cdot||_F$ is the Frobenius norm, and $I_2$ is a $2\times2$ identity matrix. %
$\mathcal{L}_{con}$ not only leads to distinguishable node assignment but also encourages the subgraph to be compact.  Take $(S^{T}AS)_{1:}$ for example, denote $a_{11},a_{12}$ the element 1,1 and the element 1,2 of $S^{T}AS$,
\begin{equation}
\begin{aligned}
a_{11} = \sum_{i,j} A_{ij}p(V_{i}\in G_{sub}|V_{i})p(V_{j}\in G_{sub}|V_{j}),\\
a_{12} = \sum_{i,j} A_{ij}p(V_{i}\in G_{sub}|V_{i})p(V_{j}\in \overline{G_{sub}}|V_{j}).\\
\end{aligned}
\end{equation}
Minimizing $\mathcal{L}_{con}$ results in $\frac{a_{11}}{a_{11}+a_{12}}\rightarrow 1$. This occurs if $V_{i}$ is in $ G_{sub}$, the elements of $\mathcal{N}(V_{i})$ have a high probability in $ G_{sub}$.  Minimizing $\mathcal{L}_{con}$ also encourages $\frac{a_{12}}{a_{11}+a_{12}}\rightarrow 0$. This encourages $p(V_{i}\in G_{sub}|V_{i})\rightarrow 0/1$ and less cuts between $G_{sub}$ and $\overline{G_{sub}}$. This also holds for $\overline{G_{sub}}$ when analyzing $a_{21}$ and $a_{22}$. \cite{zhou2004learning}

In a word, $\mathcal{L}_{con}$ encourages distinctive $S$ to stabilize the training process and a compact topology in the subgraph. Therefore, the overall loss is:
\begin{equation}
\begin{aligned}
&\min \limits_{\theta,\phi_{1}} \quad \mathcal{L}(\theta,\phi_{1},\phi_{2}^{*}) = \mathcal{L}_{cls} + \alpha \mathcal{L}_{con}+ \beta \mathcal{L}_{\rm MI} \\
&\text{ s.t. }\quad \phi_{2}^{*} = \mathop{\arg\max} \limits_{\phi_{2}}\mathcal{L}_{\rm MI}.
\end{aligned}
\label{final-bilevel}
\end{equation}

We provide the pseudo code in the Algorithm~\ref{pseudo-code} to better illustrate how to optimize the above objective.

\section{Experiments}
In this section, we evaluate the proposed SIB method on four scenarios, including improvement of graph classification, graph interpretation, graph denoising, and relevant 3D structure extraction.

\subsection{Baselines and settings}

\textbf{Improvement of graph classification:} We first examine SIB's capability of improving graph classification. Different from prior methods which aggregate all node information for the graph representation, SIB only aggregates the node message in the corresponding IB-subgraph. We plug SIB into various backbones including GCN \cite{gcn}, GAT \cite{velickovic2017graph}, GIN \cite{Xu:2019ty} and GraphSAGE \cite{conf/nips/HamiltonYL17}. That is, we first employ the backbones to extract node representations and then yield the graph representation by aggregating the node representation in the IB-Subgraph. We compare the proposed method with the mean/sum aggregation \cite{gcn,velickovic2017graph,conf/nips/HamiltonYL17,Xu:2019ty} and pooling aggregation \cite{sortpool,asapool,diffpool,journals/corr/abs-1905-10990} in terms of classification accuracy. Moreover, we apply DropEdge \cite{rong2020dropedge} to GAT, namely GAT+DropEdge, which randomly drop 30\% edges in message-passing at node-level. Similarly, we apply SIB to GAT+DropEdge, resulting in GAT+SIB+DropEdge. This is to examine the flexibility of SIB to the recent proposed dropping-based regularization in node message-passing. For fair comparisons, all the backbones for different methods consist of the same 2-layer GNN with 16 hidden-size.

\textbf{Graph Interpretation:} As the IB-subgraph preserves the predictive substructure for the graph property, it is interesting to see if we could identify the substructures which best represent some chemical properties of molecules. The goal of graph interpretation is to find the substructure which shares the most similar property to the molecule. If the substructure is disconnected, we evaluate its largest connected part. We compare SIB with the attention mechanism \cite{hierG}. That is, we attentively aggregate the node information with the normalized attention scores for graph prediction. The interpretable subgraph is generated by choosing the nodes with top $50\%$ and $70\%$ attention scores, namely Att05 and Att07. SIB outputs the interpretation with the IB-Subgraph, without a manually-selected threshold. Then, we evaluate the absolute property bias (the absolute value of the difference between the property of graph and subgraph) between the graph and its interpretation. Similarly, for fare comparisons, all the backbones for different methods consist of the same 2-layer GNN with 16 hidden-size.

\textbf{Graph Denoising:} We further evaluate the robustness of SIB to the graphs with noisy structures. We translate the permuted graph into the line-graph and use SIB and attention to infer the real structure of the graph. Then, we classify the permuted graph via the inferred structure respectively. We further compare the performance of GCN and DiffPool on the permuted graphs. Similarly, for fare comparisons, all the backbones for different methods consist of the same 2-layer GNN with 16 hidden-size.

\textbf{Relevant 3D Structure Extraction:} Most work on weakly supervised 3D segmentation still requires a  tiny  portion  of  labeled points for training \cite{xu2020weakly, shi2021label}. Here we consider a more challenging scenario. That is, to extract relevant 3D structures which belong to the same category without any node-level annotation.  This poses a huge challenge for SIB to effectively process large scale graphs. We use the GCN with residual graph connections between layers and GCN with dense graph connections which concatenates representation from previous layers in \cite{li2019deepgcns} as our backbones. After each layer, we dynamically updating the constructed K-NN graphs. We refer to these two backbones as ResDyGCN and DenseDyGCN. Since computing the connective loss in large graphs is prone to exceed the GPU memory, we employ Gumbel-Softmax strategy \cite{jang2016categorical} for continuous relaxation in learning assignment matrix.  We compare SIB with attention.  For fair comparisons, all the backbones for different methods consist of the same 4-layer model with 1024 hidden-size.

\begin{table}[t]
  \centering
  \vspace{0.2cm}
  \caption{Statistics of datasets in improvement of graph classification.}
  \setlength{\tabcolsep}{0.1mm}{
    \begin{tabular}{cccccc}
    \toprule
       & {MUTAG} & {PROTEINS} & {IMDB-BINARY} & {DD} \\
    \midrule
    Nodes & 97.9K  & 43.5K  &  19.8K &  334.9K  \\
    Edges &  202.5K & 162.1K  &  386.1K &  1.7M  \\
    Density &  4.2 $\times 10^{-5}$ & 1.7  $\times 10^{-4}$  & 2.0  $\times 10^{-3}$  &  3.0 $\times 10^{-5}$  \\
    Maximum degree   & 20   &  50 &  540  &  38  \\
    Minimum degree   &  2  &  2 &   4 &   2 \\
    Average degree   &  4  &  7 &  39  & 10   \\
    Number of triangles   & 2.8K   & 366K  & 18.8M   &  7.1M  \\   
        Maximum k-core  &  5  &  9 &  117  &  15  \\
     \makecell{Average number\\ of triangles}   & 0 & 8  &  951  & 21   \\  
    \makecell{Maximum number\\ of triangles}   &  12  &  136 &  17.8K  &   160 \\  
    \makecell{Average clustering\\ coefficient}   &  0.001965  & 0.316645  &  0.831934  & 0.413379   \\  
    \makecell{Fraction of closed\\ triangles}   &  0.003160  & 0.315106  &  0.803561  & 0.410832   \\  
    \makecell{Lower bound of\\ Maximum Clique}  &  6  & 5  &  18  &  4  \\
    \bottomrule
    \end{tabular}}%
  \label{stat:graph}%
\vspace{0.2cm}
\end{table}%

\begin{table}[t]
  \centering
  \caption{Statistics of datasets in graph interpretation.}
  \setlength{\tabcolsep}{0.2mm}{
    \begin{tabular}{ccccc}
    \toprule
    &\textbf{QED} & \textbf{DRD2} & \textbf{HLM-CLint} & \textbf{MLM-CLint} \\
    \midrule
    Number of graphs & 35000 & 3000  & 25850 & 16666 \\
    Maximum number of nodes & 29    & 66    & 37    & 37 \\
    Minimum number of nodes & 12    & 13    & 9     & 7 \\
    Average number of nodes & 21.82 & 27.43 & 25.14 & 22.44 \\
    Maximum number of edges & 34    & 74    & 42    & 42 \\
    Minimum number of edges & 12    & 14    & 9     & 7 \\
    Average number of edges & 23.43 & 30.24 & 22.23 & 24.19 \\
    Dimension of node features & 9     & 8     & 9     & 9 \\
    \bottomrule
    \end{tabular}}%
  \label{stat:interpretation}%
\end{table}%

\subsection{Datasets}

\textbf{Improvement of Graph Classification:} We evaluate different methods on the datasets of  \textbf{MUTAG} \cite{PhysRevLett.108.058301}, \textbf{PROTEINS} \cite{conf/ismb/BorgwardtOSVSK05}, \textbf{IMDB-BINARY} and \textbf{DD} \cite{nr} datasets\footnote{We follow the protocol in \url{https://github.com/rusty1s/pytorch_geometric/tree/master/benchmark/kernel}}. The statistics of the datasets are available in Table~\ref{stat:graph} \footnote{The statistics of datasets in improvement of graph classification are collected from \url{http://networkrepository.com}}.

\textbf{Graph Interpretation:} We construct the datasets for graph interpretation on four molecule properties based on ZINC dataset \cite{irwin2005zinc}], which contains 250K molecules. \textbf{QED} measures the drug likeness of a molecule, which is bounded within the range $(0,1.0)$. \textbf{DRD2} measures the probability that a molecule is active against dopamine type 2 receptor, which is bounded with $(0,1.0)$. \textbf{HLM-CLint} and \textbf{MLM-CLint} are  estimated values of in vitro human and mouse liver microsome metabolic stability (base 10 logarithm of mL/min/g). We sample the molecules with QED $ \geq 0.85$, DRD2 $ \geq 0.50$, HLM-CLint $\geq 2$,  MLM-CLint $\geq 2$ for each task. We use $85\%$ of these molecules for training, $5\%$ for validating, and $10\%$ for testing. The statistics of the datasets are available in Table~\ref{stat:interpretation}.

\textbf{Graph Denoising:} We generate a synthetic dataset by adding $30\%$ redundant edges for each graph in \textbf{MUTAG} dataset. We use $70\%$ of these graphs for training, $5\%$ for validating, and $25\%$ for testing. 

\begin{figure}[t]
\centerline{\includegraphics[width=1.0\columnwidth]{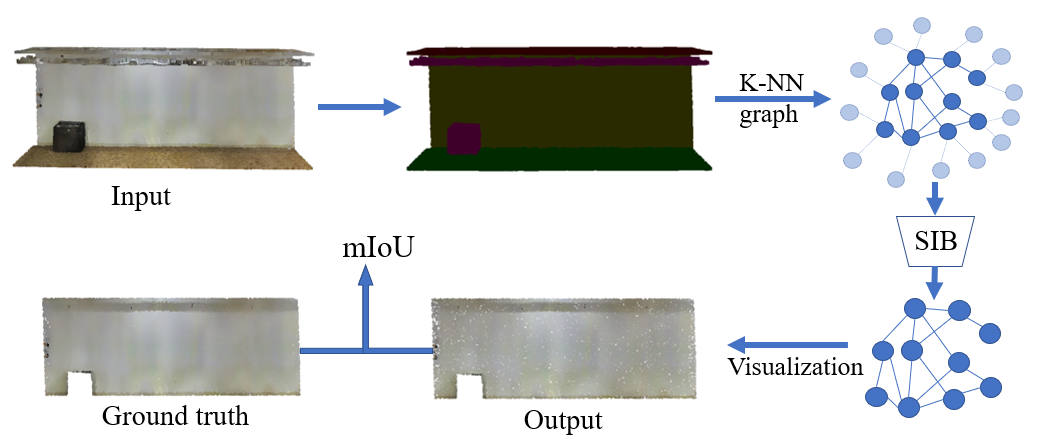}}
\caption{The pipeline of relevant 3D structure extraction. We label the point cloud with the category of its main component and leverage the color and coordinates as the feature of each point. Then we convert the point cloud into a K-NN graph, where K equals 16. SIB further takes graph data as input and extracts the structure which is most relevant to its main component. We evaluate the result by computing the mIoU of the output and the ground truth (not available in the training procedure) in the testing set. This example is from Area 5 in S3DIS dataset.}
\label{pipeline-point}
\end{figure}

\textbf{Relevant 3D Structure Extraction:} We construct the datasets for relevant 3D Structure Extraction based on S3DIS dataset \cite{armeni_cvpr16}. It contains over 23000 point clouds of 6 indoor areas. Every point cloud consists of 4096 points. Each point is labeled with one of 13 categories, and it has a 9-dimensional feature which indicates the 3D coordinates, normalized 3D coordinates, and RGB color. To enable SIB training, we first choose the point clouds with the percent of main categories greater than 50\%, and label the point clouds with the main categories. Then we choose \textbf{floor} and \textbf{wall} for relevant structure extraction as these two categories contain sufficient training samples. We use Area 1-5 for training and Area 6 for testing, which yield 637 and 2099 training samples and 411 and 1270 testing samples for floor and wall respectively. We train these two categories separately and randomly add the same number of negative samples from the rest 12 categories in training. All point clouds are transformed into K-NN graphs, where K equals 16.

\begin{table}[t]
  \centering
  \vspace{0.1cm}
  \caption{Quantitative results on graph denoising. We report the classification accuracy (Acc), the number of real edges over total real edges (Recall), and the number of real edges over total edges in subgraphs (Precision) on the test set.}
  \setlength{\tabcolsep}{0.8mm}{
    \begin{tabular}{cccccc}
    \toprule
    Method & {GCN} & {DiffPool} & {GCN+Att05} & {GCN+Att07} & \textbf{GCN+SIB} \\
    \midrule
    Recall & - & - & {0.226$\pm$0.047} & {0.324$\pm$ 0.049} & \textbf{0.493$\pm$ 0.035} \\
    Precision & - & - & {0.638$\pm$ 0.141} & {0.675$\pm$ 0.104} & \textbf{0.692 $\pm$0.061} \\
    Acc   & 0.617 & 0.658 & 0.649 & 0.667 & \textbf{0.684} \\
    \bottomrule
    \end{tabular}}%
  \label{tab:edge}%
\vspace{0.1cm}
\end{table}%

\begin{figure*}[t]
\begin{center}
\centerline{\includegraphics[width=1.9\columnwidth]{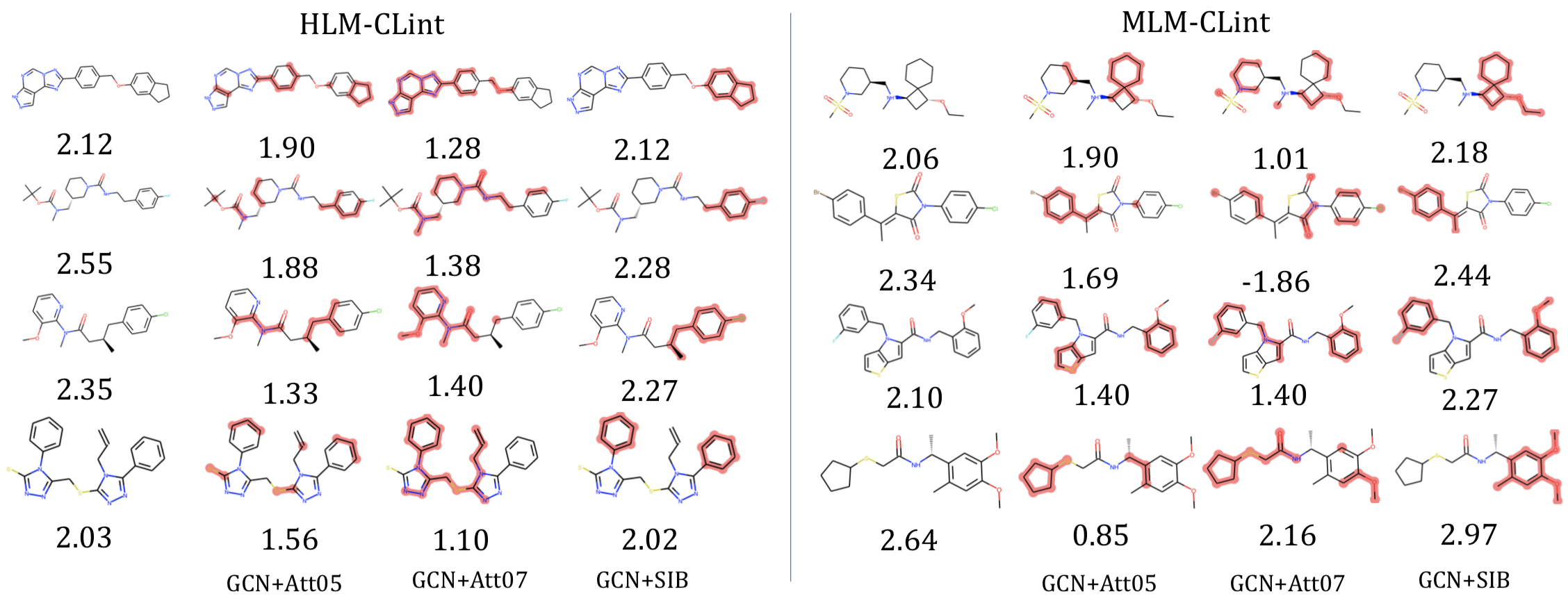}}
\end{center}
\vspace{0.2cm}
\caption{The molecules with their interpretable subgraphs discovered by different methods. These subgraphs exhibit similar chemical properties compared to the molecules on the left.}
\label{mol-compare}
  \vspace{0.2cm}
\end{figure*}

\subsection{Results}

\subsubsection{Improvement of Graph Classification}

In Table \ref{tab:2}, we comprehensively evaluate the proposed method and baselines on improvement of graph classification. We train SIB on various backbones and aggregate the graph representations only from the IB-subgraphs. As is shown, SIB outperforms the backbones by a large margin. This is because SIB can recognize the predictive yet compressed substructures and thus relieves the side-effect of noisy and redundant structures in graph data, which is detrimental to graph classification. SIB also exceeds many powerful pooling-based methods, which is able to leverage the hierarchical topological information in graphs. Notice that SIB operates on node features rather than hierarchically coarsened features, therefore, binding SIB with pooling-based methods may be problematic. However, it is sufficient to say that SIB can improve graph classification with the information theoretic IB-subgraph.

Recent work shows that regularization on message-passing can lead to better node representation \cite{rong2020dropedge,zheng2020robust}.  This is orthogonal to our work as these methods either relieves over-smoothing in GCN or select representative neighborhoods on node-level tasks. Our SIB recognizes a predictive yet compressed subgraph, namely IB-subgraph, to enhance the performance of existing models on graph-level tasks. As SIB also relies on informative node representations, we further plug DropEdge into our model and also obtain a significant gain in performance.

\begin{table}
  \centering
  \vspace{0.2pt}
  \caption{Average number of disconnected substructures per graph selected by different methods.}
    \begin{tabular}{ccccc}
    \toprule
    Method & QED   & DRD2  & HLM   & MLM \\
    \midrule
    GCN+Att05 & 3.38 & 1.94 & 3.11 & 5.16 \\
    GCN+Att07 & 2.04 & 1.76 & 2.75 & 3.00 \\
    \textbf{GCN+SIB}   & \textbf{1.57} & \textbf{1.08} & \textbf{2.29} & \textbf{2.06} \\
    \bottomrule
    \end{tabular}%
  \label{tab:disconnected}%
  \vspace{0.2pt}
\end{table}

\subsubsection{Graph Interpertation}
We further examine the capability of IB-subgraph on interpreting the property of molecules. To this end, we first extract the substructures which mostly affect the chemical property of molecules by different models. Then we compare the property of learned substructures with the input molecules. To ensure the chemical validity, when the substructure is disconnected, we evaluate the property of its largest connected part \footnote{We obtain QED and DRD2 values of molecules with the toolkit on \url{https://www.rdkit.org/}. And we evaluate HLM-CLint and MLM-CLint value of molecules on \url{https://drug.ai.tencent.com/}}.

\begin{figure}[t]
\centerline{\includegraphics[width=1.0\columnwidth]{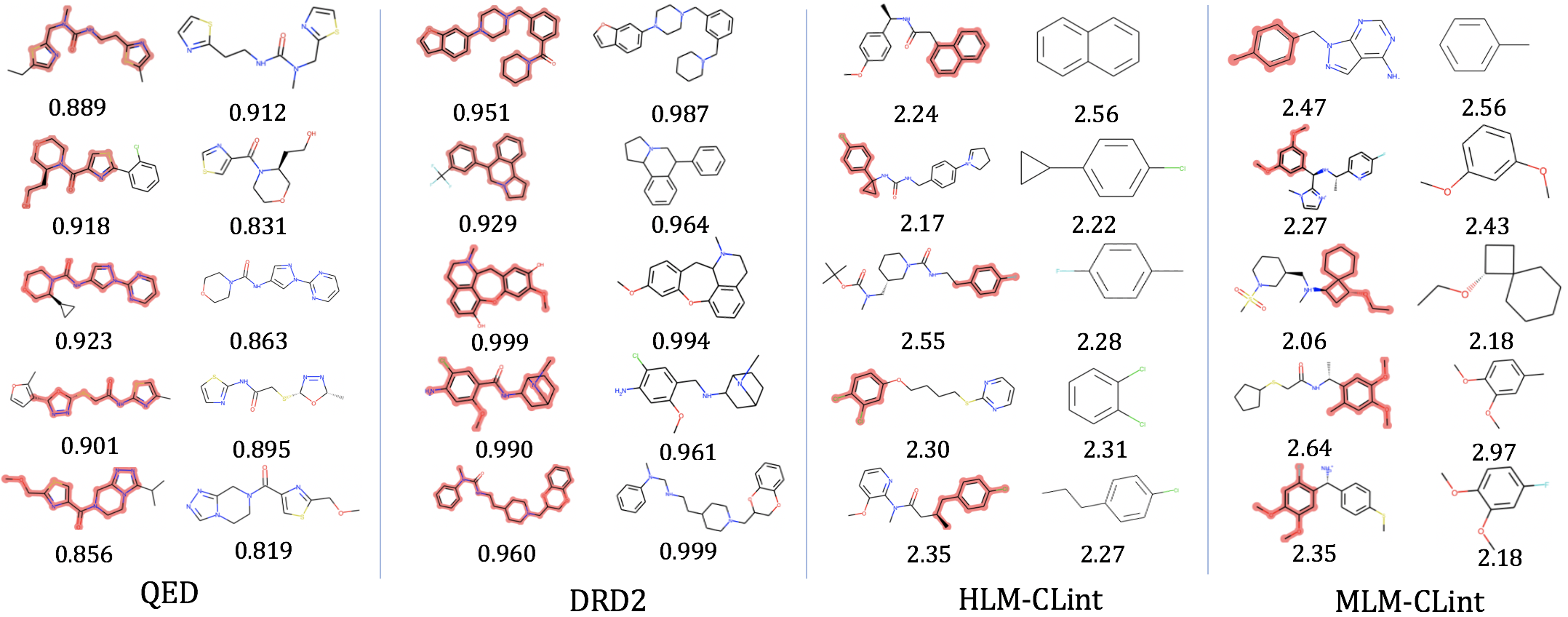}}
\caption{The molecules with its interpretation found by SIB. These subgraphs exhibit similar chemical properties compared to the molecules on the left.}
\label{fig1}
\end{figure}

Table \ref{tab:1} shows the quantitative performance of different methods on the graph interpretation task. SIB is able to generate precise graph interpretation (IB-Subgraph), as the substructures found by SIB have the most similar property to the input molecules. Moreover, it is noticed that the interpretation found by attention-based method is highly influenced by the manually selected threshold. For example, GCN+Att07 generates subgraphs with more similar properties compared to GCN+att05. However, GCN+Att05 outperforms GCN+Att07 on HLM-CLint and MLM-CLint properties. Therefore, one needs to carefully choose the threshold for different tasks. In contrast, SIB does require a threshold to select the interpretation thanks to the information theoretic objective. In Fig.~\ref{mol-compare}, SIB generates a more compact and reasonable interpretation of the property of molecules confirmed by chemical experts. More visualization results are provided in Fig.~\ref{fig1}

We further compare the compactness of the found subgraphs of different methods as a more compact subgraph has a higher probability to be a functional group and easier for chemists to interpret the property of a molecule. In Table \ref{tab:disconnected}, we compare the average number of disconnected substructures per graph. SIB generates more compact subgraphs to better interpret the graph property. Moreover, compared to the baselines, SIB does not require a hyper-parameter to control the sizes of subgraphs, thus being more adaptive to different tasks. 

\begin{figure}[t]
\begin{center}
\centerline{\includegraphics[width=1.0\columnwidth]{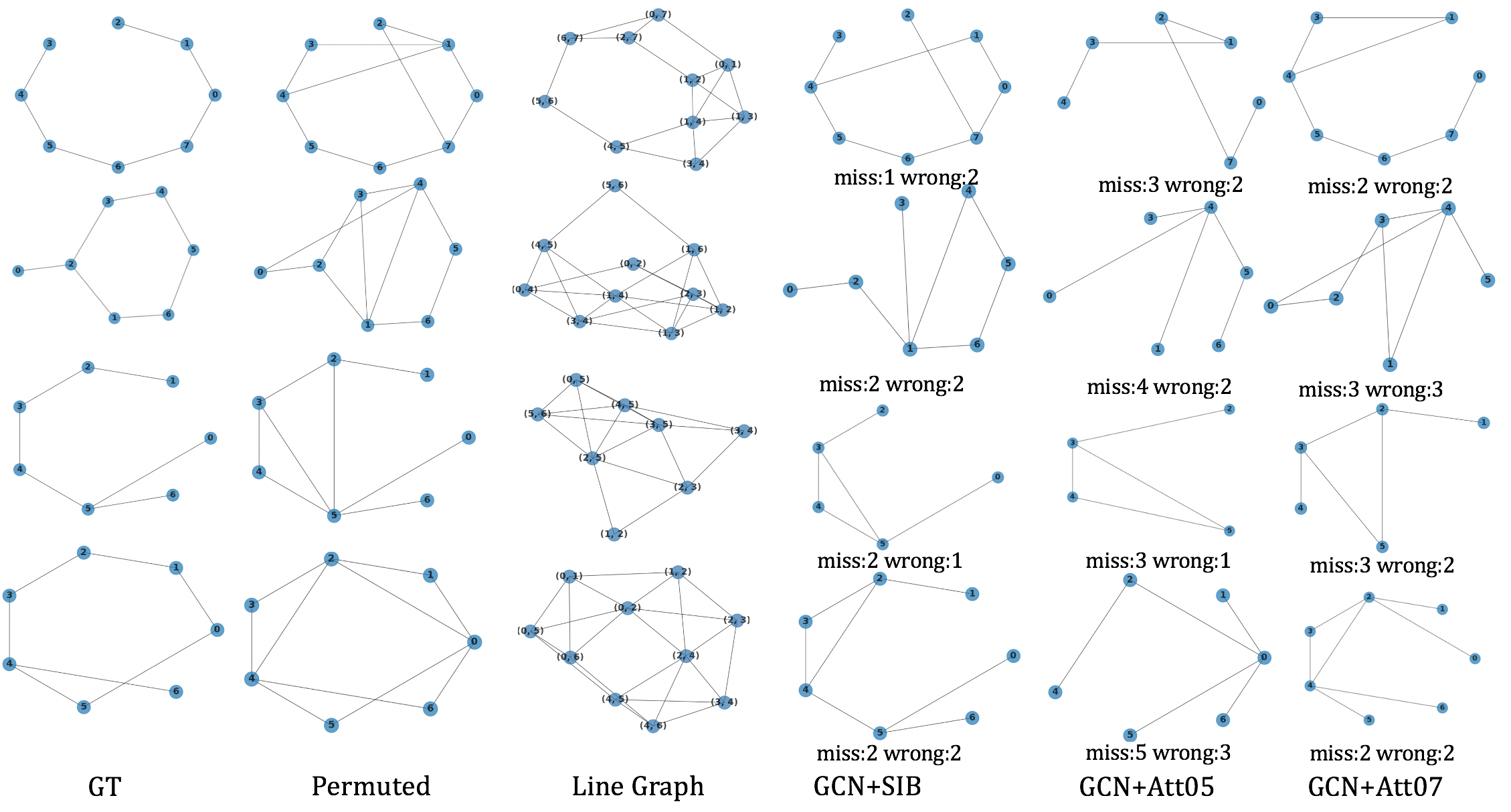}}
\end{center}
\caption{We show the blindly denoising results on permuted graphs. Each method operates on the line-graphs and tries to recover the true topology by removing the redundant edges. 
Columns 4,5,6 shows results obtained by different methods, where ``miss: $m$, wrong: $n$'' means missing $m$ edges and there are $n$ wrong edges in the output graph.
SIB always recognizes more similar structure to the ground truth (not provided in the training process) than other methods. }
\label{newedge}
\end{figure}

\subsubsection{Graph Denoising} 
Table \ref{tab:edge} shows the performance of different methods on noisy graph classification. GCN and DiffPool are vulnerable to structure perturbation since they are unable to distinguish the real structure and  noise. However, SIB is able to better reveal the real structure of permuted graphs in terms of precision and recall rate of true edges, even in the absence of explicit annotations. Therefore, SIB is more robust to perturbations and outperforms the baselines on classification accuracy by a large margin. We provide visualization results in Fig.~\ref{newedge}. It is noticed that SIB recognizes more similar structures to the ground truth (not provided in the training process) than other methods.

\subsubsection{Relevant 3D Structure Extraction}

This task is rather difficult since no point label is provided in the scene point cloud. To the best of our knowledge, few prior methods focus on this extreme scenario. We evaluate different methods in terms of mean intersection over union (mIoU) between the learned substructure and the ground truth\cite{xu2020weakly, shi2021label}. We employ two backbones, namely ResDyGCN and DenseDyGCN, in \cite{li2019deepgcns} to extensively compare different methods. As is shown in Table \ref{3D}, SIB also outperforms the attentive methods since it is capable of effectively recognize relevant structures in large scale point clouds. The results of attentive methods are highly influenced by the threshold. We also notice using ResDyGCN as backbone leads to better results. This is because DenseDyGCN concatenates all previous point representation at each layer, which potentially brings superfluous information. Meanwhile, the residual connection between layers avoids gradient vanishing, and thus leads to more stable training and better performance. 

\begin{table}[htbp]
  \centering
  \caption{We compare the capability of different methods on extraction label-relevant substructures in 3D point cloud. Notice that no point label is available in the training process.}
    \begin{tabular}{c|l|r|r|r|r}
    \toprule
    \multirow{2}[4]{*}{Backbone} & \multicolumn{1}{c|}{\multirow{2}[4]{*}{Model}} & \multicolumn{2}{c}{floor} & \multicolumn{2}{c}{wall} \\
\cmidrule{3-6}          &       & \multicolumn{1}{l}{mIoU} & \multicolumn{1}{l|}{size} & \multicolumn{1}{l}{mIoU} & \multicolumn{1}{l}{size} \\
    \midrule
    \multirow{3}[2]{*}{ResDyGCN} & Att05 & 0.46  & 2048  & 0.384 & 2048 \\
          & Att07 & 0.464 & 2867  & 0.465 & 2867 \\
          & SIB   & 0.519 & 2819.628 & 0.492 & 2032.613 \\
    \midrule
    \multirow{3}[2]{*}{DenseDyGCN} & Att05 & 0.27  & 2048  & 0.289 & 2048 \\
          & Att07 & 0.378 & 2867  & 0.358 & 2867 \\
          & SIB   & 0.397 & 1978.479 & 0.419 & 1784.214 \\
    \bottomrule
    \end{tabular}%
  \label{3D}%
\end{table}%

\begin{table}[t]
\vspace{0.2cm}
  \centering
  \caption{Ablation study on $\mathcal{L}_{con}$ and $\mathcal{L}_{MI}$. Note that we try several initiations for SIB w/o $\mathcal{L}_{con}$ and $\mathcal{L}_{MI}$ to get the current results due to the instability of optimization process.}
  \setlength{\tabcolsep}{1mm}{
    \begin{tabular}{ccccc}
    \toprule
    \textbf{Method}    & \textbf{QED} & \textbf{DRD2}  & \textbf{HLM-CLint}   & \textbf{MLM-CLint} \\
    \midrule
    SIB w/o $\mathcal{L}_{con}$ & 0.46$\pm$ 0.07 & 0.15$\pm$ 0.12 & 0.45$\pm$ 0.37 & 1.58$\pm$ 0.86 \\
    SIB w/o $\mathcal{L}_{MI}$ & 0.43$\pm$ 0.15 & 0.21$\pm$ 0.13 & 0.48$\pm$ 0.34 & 1.20$\pm$ 0.97 \\
    \textbf{SIB}  & \textbf{0.38$\pm$ 0.12} & \textbf{0.06$\pm$ 0.09} & \textbf{0.37$\pm$ 0.30} & \textbf{0.72$\pm$ 0.55} \\
    \bottomrule
    \end{tabular}}%
  \label{ablation}%
\end{table}%

\subsubsection{Ablation Study}

To further understand the roles of $\mathcal{L}_{con}$ and $\mathcal{L}_{MI}$, we derive two variants of our method by deleting $\mathcal{L}_{con}$ and $\mathcal{L}_{MI}$, namely SIB w/o $\mathcal{L}_{con}$ and SIB w/o $\mathcal{L}_{MI}$. Note that SIB w/o $\mathcal{L}_{MI}$ is similar to InfoGraph \cite{sun2019infograph} and GNNExplainer \cite{gnnexplainer}, as they only consider maximizing MI between latent embedding and global summarization and ignore compression. When adapted to subgraph recognition, it is likely to be $G=G_{sub}$. We evaluate the variants with 2-layer GCN and 16 hidden sizes on graph interpretation. In practice, we find that the training process of SIB w/o $\mathcal{L}_{con}$ is unstable as discussed in Section 4.3. Moreover, we find that SIB w/o $\mathcal{L}_{MI}$ is very likely to output $G_{sub}=G$, as it does not consider compression. Therefore, we try several initiations for SIB w/o $\mathcal{L}_{con}$ and $\mathcal{L}_{MI}$ to get the current results. As shown in Table \ref{ablation}, SIB also outperforms the variants, and thus indicates that every part of our model does contribute to the improvement of performance. 

\subsubsection{More Discussions}

\begin{table}[t]
\vspace{0.1cm}
  \centering
  \caption{The influence of the hyper-parameter $\alpha$ of $L_{con}$  to the size of subgraphs.}
    \begin{tabular}{ccccc}
    \toprule
    $\alpha$ & \multicolumn{1}{c}{1} & \multicolumn{1}{c}{3} & \multicolumn{1}{c}{5} & \multicolumn{1}{c}{10} \\
    \midrule
    All   & 0.483$\pm$0.143 & 0.496$\pm$0.150 & 0.494$\pm$0.147 & 0.466$\pm$0.150 \\
    Max   & 0.387$\pm$0.173 & 0.413$\pm$0.169 & 0.411$\pm$0.169 & 0.391$\pm$0.172 \\
    \bottomrule
    \end{tabular}%
  \label{sizeofsubgraph}%
  \vspace{0.1cm}
\end{table}%

\begin{table}
\vspace{0.1cm}
  \centering
  \caption{The overlap between the chosen subgraphs with different initialization.}
  \setlength{\tabcolsep}{0.01mm}{
    \begin{tabular}{@{ }c|@{ }c|@{ }c|@{ }c|@{ }c|@{ }c@{ }}
    \toprule
     Run     & 1    & 2    & 3    & 4    & 5 \\
              \midrule
    $IoU_{all}$ & 0.848$\pm$0.163 & 0.765$\pm$0.106 & 0.784$\pm$0.112 & 0.829$\pm$0.166 & 0.813$\pm$0.186 \\
    $IoU_{max}$ & 0.779$\pm$0.330 & 0.696$\pm$0.310 & 0.742$\pm$0.333 & 0.757$\pm$0.335 & 0.762$\pm$0.304 \\
    \bottomrule
    \end{tabular}}%
  \label{init}
  \vspace{0.1cm}
\end{table}%

$L_{con}$ is proposed for stabilizing the training process and resulting in compact subgraphs. As it poses regularization for the subgraph generation, we are interested in its potential influence on the sizes of the chosen IB-Subgraphs. Therefore, we show the influence of different hyper-parameters of $L_{con}$ to the sizes of the chosen IB-Subgraphs. We implement the experiments with $\alpha$ varies in $\{1,3,5,10\}$ on QED dataset and compute the mean and deviation of the sizes of IB-Subgraphs (All) and their largest connected parts (Max). As shown in Table \ref{sizeofsubgraph}, we observe that different values of $\alpha$ result in similar sizes of IB-Subgraphs. Therefore, its influence on the size of chosen subgraphs is weak.

As the initialization of our model may potentially influence the final chosen subgraphs, we rerun our model five times on the QED dataset for graph interpretation task. Then, we employ the intersection over union ($IoU$) to measure the overlap between the subgraphs in 5 different runs and the results reported in Table \ref{tab:1}. Similarly, we compute the $IoU$ between the chosen subgraphs and their largest connected parts separately, which refer to $IoU_{all}$ and $IoU_{max}$. We finally report the mean and standard deviation of $IoU_{all}$, $IoU_{max}$ on the testing set in Table \ref{init}. We notice that different initialization has limited influence on the chosen subgraphs, as all the results of five additional runs have high portions of common nodes with the initial run. 

\begin{table}[t]
  \centering
  \caption{Size of the chosen subgraphs on four datasets in percent.}
    \begin{tabular}{ccccc}
    \toprule
    Method & \textbf{QED} & \textbf{DRD2} & \textbf{HLM-CLint} & \textbf{MLM-CLint} \\
    \midrule
    GCN+Att05 & 43.5$\pm$5.4 & 46.8$\pm$3.1 & 48.1$\pm$4.1 & 45.1$\pm$4.1 \\
    GCN+Att07 & 65.8$\pm$3.4 & 66.7$\pm$3.0 & 65.8$\pm$5.7 & 67.6$\pm$5.5 \\
    GCN+SIB   & 49.6$\pm$15.0 & 94.8$\pm$5.3 & 47.7$\pm$13.7 & 54.7$\pm$17.2 \\
    \bottomrule
    \end{tabular}%
  \label{allscale}%
\end{table}%

\begin{table}[t]
  \centering
  \caption{Size of largest connected parts used for graph interpretation in percent.}
    \begin{tabular}{ccccc}
    \toprule
    Method & \textbf{QED} & \textbf{DRD2} & \textbf{HLM-CLint} & \textbf{MLM-CLint} \\
    \midrule
    GCN+Att05 & 22.5$\pm$9.5 & 34.7$\pm$9.4 & 23.5$\pm$7.2 & 29.7$\pm$7.9 \\
    GCN+Att07 & 43.3$\pm$11.8 & 54.2$\pm$13.0 & 45.0$\pm$15.2 & 41.2$\pm$8.4 \\
    GCN+SIB   & 41.3$\pm$16.9 & 92.8$\pm$10.4 & 29.1$\pm$10.6 & 36.9$\pm$16.2 \\
    \bottomrule
    \end{tabular}%
  \label{maxscale}%
\end{table}%


In the graph interpretation task, the hyper-parameter of $L_{con}$, $\alpha$, is set to be 5 on four datasets. We show the mean and standard deviation of the sizes of subgraphs in percent in Table \ref{allscale} and Table \ref{maxscale}. Note that the sizes of chosen subgraphs  mainly depend on task relevant information. For example, as DRD2 measures the probability of being active against dopamine type 2 receptor, it depends on almost the whole structure of a molecule. In contrast, HLM-CLint measures vitro human microsome metabolic stability, which is greatly influenced by small motifs. As shown in Table \ref{allscale} and Table \ref{maxscale}, GCN+SIB can recognize the subgraphs with adaptive sizes on different tasks, leading to better performance. However, in GCN+Att05 and GCN+Att07,  the size of subgraphs is explicitly controlled by the hyper-parameter (preserve top 50\% or 70 \% nodes with the highest attention scores). Therefore, the performances of these methods are limited.

\section{Conclusion}

In this paper, we have studied a subgraph recognition problem to extract a predictive yet compressed subgraph, termed the IB-subgraph. To effectively recognize the IB-subgraph in a weakly supervised manner, we propose a novel subgraph information bottleneck (SIB) framework. Unlike the prior work in information bottleneck, SIB directly recognizes the predictive IB-subgraph for the graph label and operates on the irregular graph-structured data. We further propose a bi-level scheme to efficiently optimize SIB with a mutual information estimator for irregular graph data. We introduce continuous relaxation to enable training SIB with the gradient-based optimizer and a connectivity loss to stabilize the training process. We evaluate the model-agnostic SIB framework on both graph learning and computer vision scenarios, including the improvement of graph classification, graph interpretation, graph denoising, and relevant 3D structure extraction. Experiment results verify the superior properties of IB-subgraphs.

\appendices

\ifCLASSOPTIONcompsoc
\else
\fi


\ifCLASSOPTIONcaptionsoff
  \newpage
\fi



\bibliographystyle{IEEEtran}
\bibliography{tpami}
%



%


\begin{IEEEbiography}[{\includegraphics[width=1in,height=1.25in,clip,keepaspectratio]{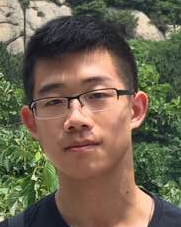}}]{Junchi Yu}
Junchi Yu is an M.S. student at the Institute of Automation, Chinese Academy of Sciences, Beijing, China. He obtained the B.E. degree from Wuhan University, Wuhan, China. His main research interests include generative models, graph neural networks, and graph generations. He has published several papers on machine learning top conference ICLR, IJCAI, etc.
\end{IEEEbiography}

\begin{IEEEbiography}[{\includegraphics[width=1in,height=1.25in,clip,keepaspectratio]{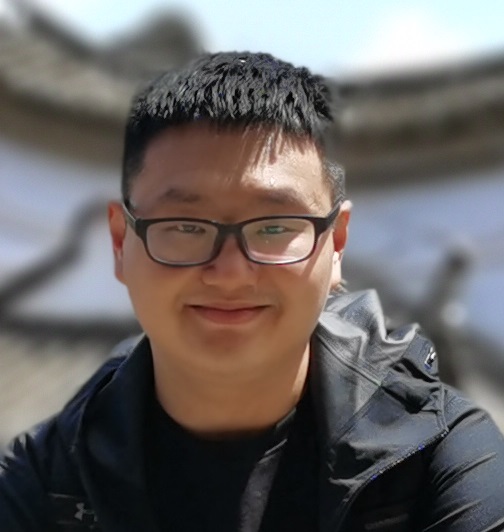}}]{Tingyang Xu} is a Senior researcher of Machine Learning Center in Tencent AI Lab. He obtained the Ph.D. degree from The University of Connecticut in 2017 and joined Tencent AI Lab in July 2017. In Tencent AI Lab, he is working on deep graph learning, graph generations and applying the deep graph learning model to various applications, such as molecular generation and rumor detection. His main research interests include social network analysis, graph neural networks, and graph generations, with particular focus on design deep and complex graph learning models for molecular generations. He has published several papers on data mining, machine learning top conferences KDD, WWW, NeurIPS, ICLR, CVPR, ICML, etc.
\end{IEEEbiography}

\begin{IEEEbiography}[{\includegraphics[width=1in,height=1.25in,clip,keepaspectratio]{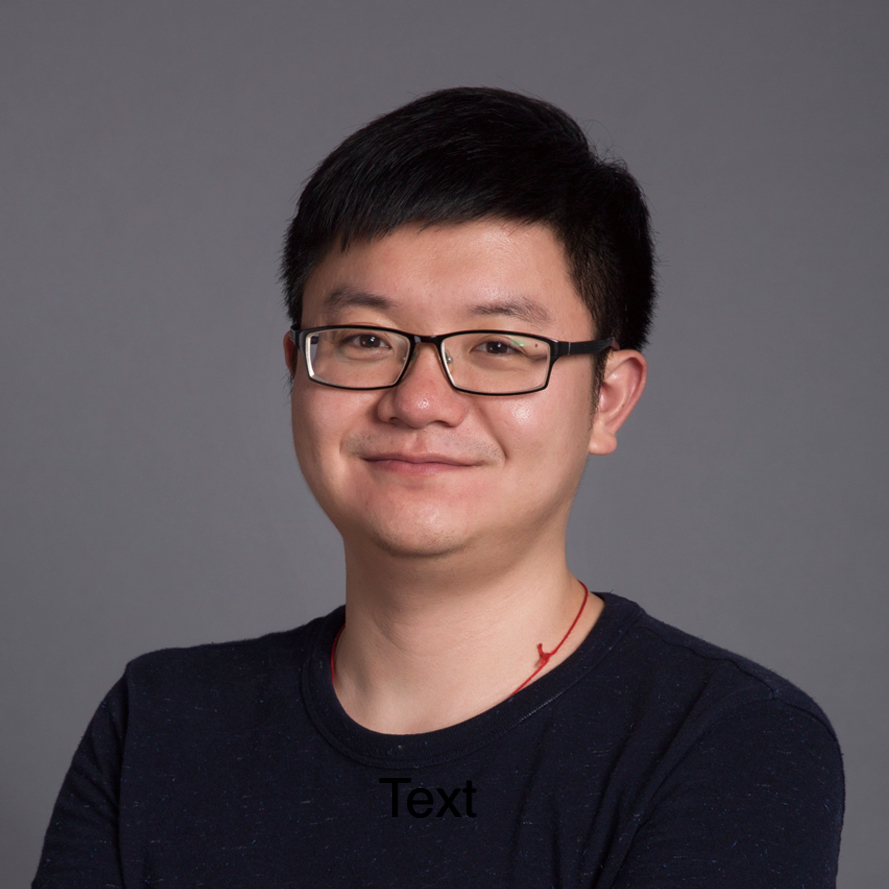}}]{Yu Rong} is a Senior researcher of Machine Learning Center in Tencent AI Lab. He received the Ph.D. degree from The Chinese University of Hong Kong in 2016. He joined Tencent AI Lab in June 2017.  His main research interests include social network analysis, graph neural networks, and large-scale graph systems. In Tencent AI Lab, he is working on building the large-scale graph learning framework and applying the deep graph learning model to various applications, such as ADMET prediction and malicious detection. He has published several papers on data mining, machine learning top conferences, including the Proceedings of KDD, WWW, NeurIPS, ICLR, CVPR, ICCV, etc. 
\end{IEEEbiography}

\begin{IEEEbiography}[{\includegraphics[width=1in,height=1.25in,clip,keepaspectratio]{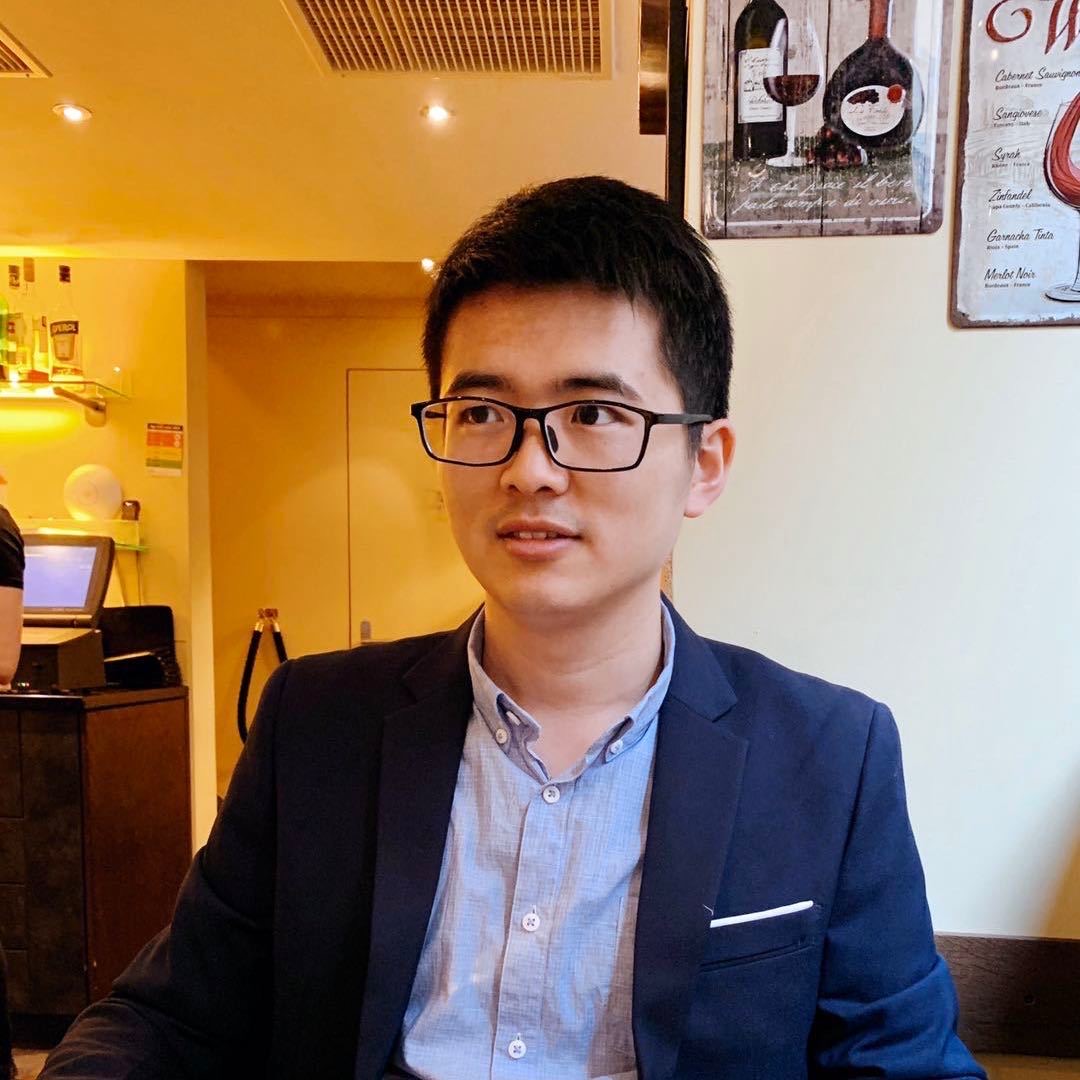}}]{Yatao Bian}is a Senior researcher of Machine Learning Center in Tencent AI Lab. He received the Ph.D. degree from the Institute for Machine Learning at ETH Zurich and joined Tencent AI Lab in February 2020. He has been an associated Fellow of the Max Planck ETH Center for Learning Systems since June 2015. Before the Ph.D. program, he obtained both of his M.Sc.Eng. and B.Sc.Eng. degrees from Shanghai Jiao Tong University. He is now working on graph neural networks, optimization for machine learning and applications such as structure-based drug discovery, 3D protein modeling and social network analysis. He has won the National Champion in AMD China Accelerated Computing Contest 2011-2012. 
\end{IEEEbiography}

\begin{IEEEbiography}[{\includegraphics[width=1in,height=1.25in,clip,keepaspectratio]{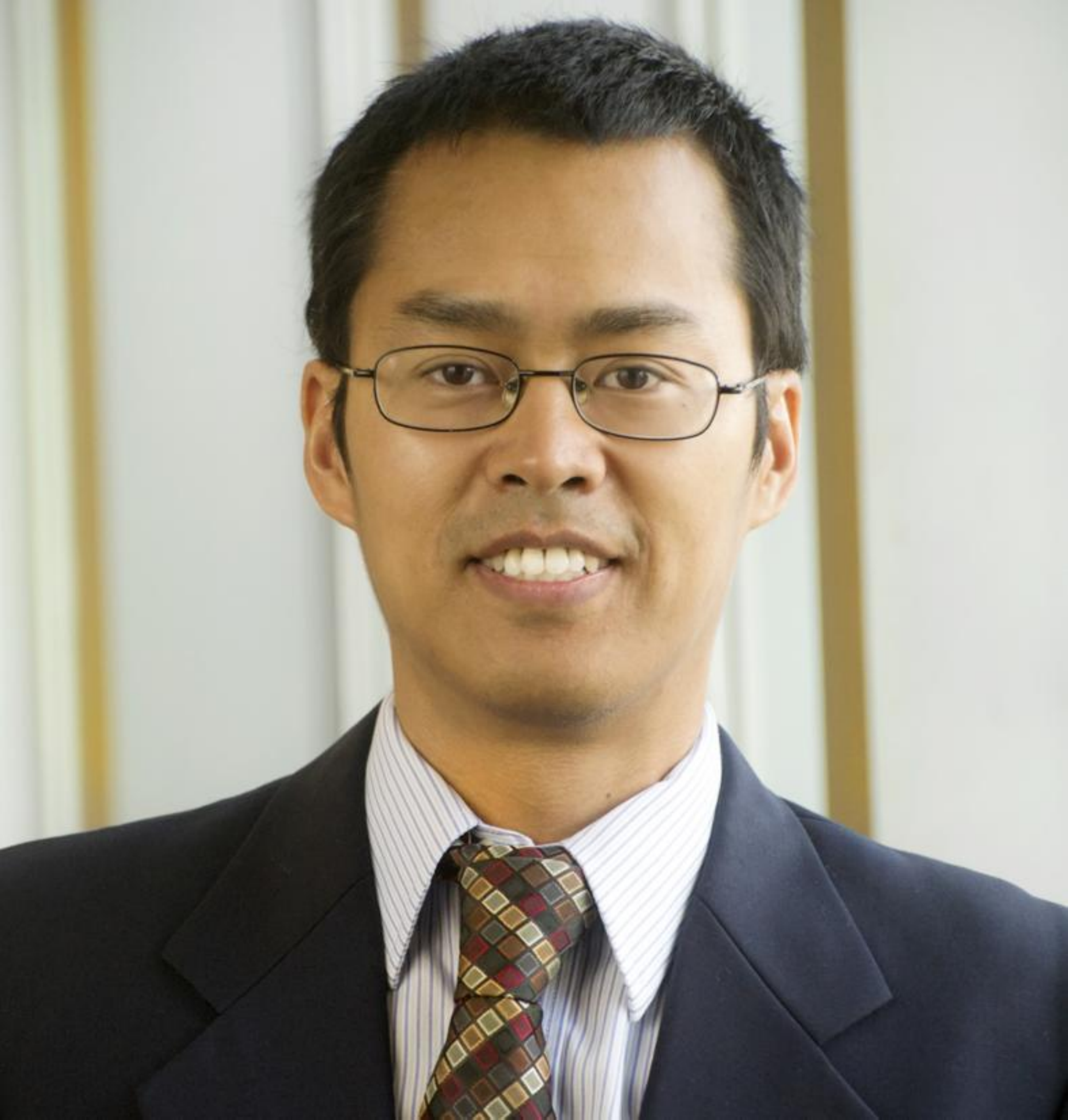}}]{Junzhou Huang}
is an Associate Professor in the Computer Science and Engineering department at the University of Texas at Arlington. He also served as the director of machine learning center in Tencent AI Lab. He received the B.E. degree from Huazhong University of Science and Technology, Wuhan, China, the M.S. degree from the Institute of Automation, Chinese Academy of Sciences, Beijing, China, and the Ph.D. degree in Computer Science at Rutgers, The State University of New Jersey. His major research interests include machine learning, computer vision and imaging informatics. He was selected as one of the 10 emerging leaders in multimedia and signal processing by the IBM T.J. Watson Research Center in 2010. His work won the MICCAI Young Scientist Award 2010, the FIMH Best Paper Award 2011, the MICCAI Young Scientist Award Finalist 2011, the STMI Best Paper Award 2012, the NIPS Best Reviewer Award 2013, the MICCAI Best Student Paper Award Finalist 2014 and the MICCAI Best Student Paper Award 2015. He received the NSF CAREER Award in 2016. 
\end{IEEEbiography}

\begin{IEEEbiography}[{\includegraphics[width=1in,height=1.25in,clip,keepaspectratio]{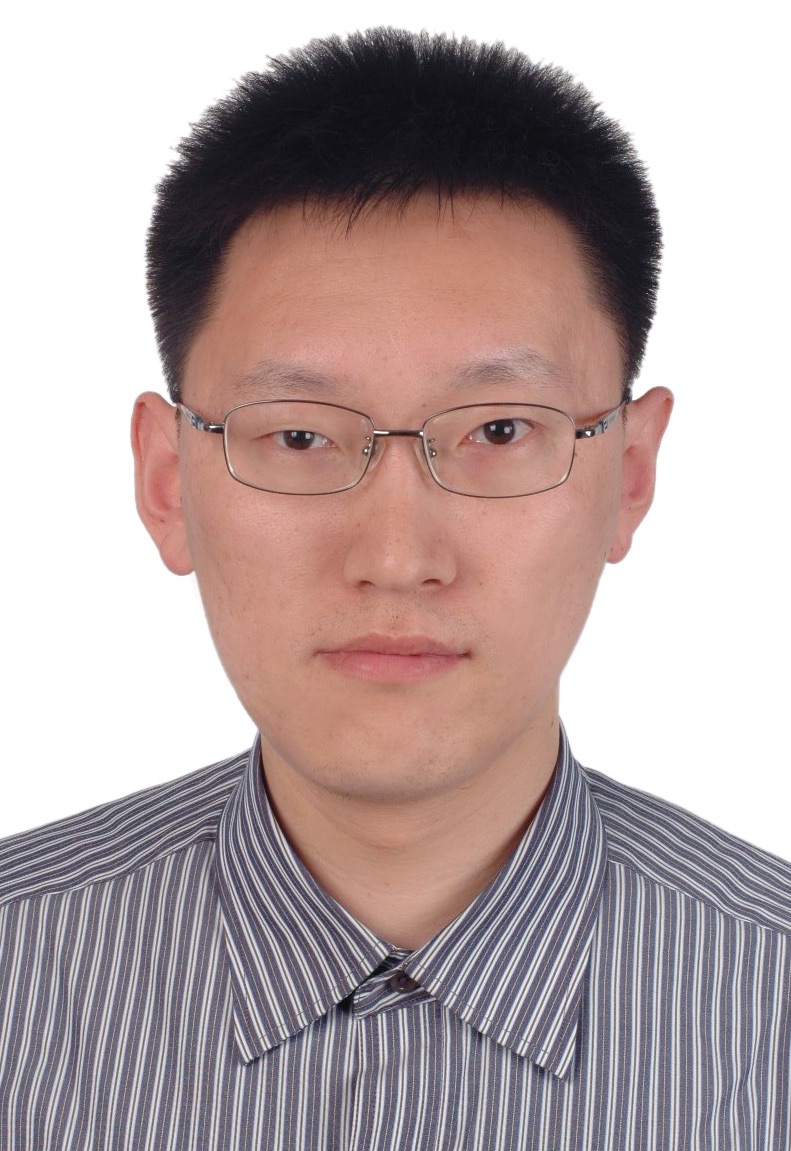}}]{Ran He}
received the B.E. degree in Computer Science from Dalian University of Technology, the M.S. degree in Computer Science from Dalian University of Technology, and Ph.D. degree in Pattern Recognition and Intelligent Systems from CASIA in 2001, 2004 and 2009, respectively. Since September 2010, Dr. He has joined NLPR where he is currently a full Professor. He serves as an associate editor of Pattern Recognition (Elsevier), and serves on the program committee of several conferences. He is a fellow of IAPR. His research interests focus on information theoretic learning, pattern recognition and computer vision.
\end{IEEEbiography}



\end{document}